\documentclass[10pt,journal,compsoc]{IEEEtran}
\ifCLASSOPTIONcompsoc
  \usepackage[nocompress]{cite}
\else
  \usepackage{cite}
\fi
\ifCLASSINFOpdf
\else
\fi

\usepackage{algorithm, algorithmic}
\usepackage{multirow}
\usepackage{setspace}
\usepackage{amsmath}
\usepackage{amssymb}
\usepackage{bm}
\usepackage{booktabs}
\usepackage{threeparttable}

\usepackage{graphicx}
\usepackage{multirow}
\usepackage{bbding}
\usepackage{subfig}

\usepackage[pagebackref=true,breaklinks=true,letterpaper=true,colorlinks,bookmarks=false]{hyperref}

\begin{document}
%
\title{Reusable Architecture Growth \\ for Continual Stereo Matching}
%
%
%
%

\author{Chenghao~Zhang,
        Gaofeng~Meng,~\IEEEmembership{Senior Member,~IEEE,}
        Bin~Fan,~\IEEEmembership{Senior Member,~IEEE,}
        Kun~Tian,
        Zhaoxiang~Zhang,~\IEEEmembership{Senior Member,~IEEE,}
        Shiming~Xiang,
        and~Chunhong~Pan
\IEEEcompsocitemizethanks{\IEEEcompsocthanksitem C.~Zhang, G.~Meng, K.~Tian, Z.~Zhang, S.~Xiang, and C.~Pan are with the State Key Laboratory of Multimodal Artificial Intelligence Systems, Institute of Automation, Chinese Academy of Sciences, Beijing 100190, China.
E-mail: \{chenghao.zhang, gfmeng, kun.tian, zxzhang, smxiang, chpan\}@nlpr.ia.ac.cn.
\IEEEcompsocthanksitem C.~Zhang, G.~Meng, K.~Tian, Z.~Zhang, and S.~Xiang are also with the School of Artificial Intelligence, University of Chinese Academy of Sciences, Beijing 100049, China.
\IEEEcompsocthanksitem  G.~Meng and Z.~Zhang are also with the Centre for Artificial Intelligence and Robotics, HK Institute of Science \& Innovation, Chinese Academy of Sciences.
\IEEEcompsocthanksitem  B.~Fan is with the School of Intelligence Science and Technology and the Institute of Artificial Intelligence, University of Science and Technology Beijing, Beijing 100083, China. E-mail: bin.fan@ieee.org.
\IEEEcompsocthanksitem  G.~Meng and B.~Fan are the corresponding authors.}
}

\IEEEtitleabstractindextext{%
\begin{abstract}
The remarkable performance of recent stereo depth estimation models benefits from the successful use of convolutional neural networks to regress dense disparity. Akin to most tasks, this needs gathering training data that covers a number of heterogeneous scenes at deployment time. However, training samples are typically acquired continuously in practical applications, making the capability to learn new scenes continually even more crucial. For this purpose, we propose to perform continual stereo matching where a model is tasked to 1) continually learn new scenes, 2) overcome forgetting previously learned scenes, and 3) continuously predict disparities at inference. We achieve this goal by introducing a Reusable Architecture Growth (RAG) framework. RAG leverages task-specific neural unit search and architecture growth to learn new scenes continually in both supervised and self-supervised manners. It can maintain high reusability during growth by reusing previous units while obtaining good performance. Additionally, we present a Scene Router module to adaptively select the scene-specific architecture path at inference. Comprehensive experiments on numerous datasets show that our framework performs impressively in various weather, road, and city circumstances and surpasses the state-of-the-art methods in more challenging cross-dataset settings. Further experiments also demonstrate the adaptability of our method to unseen scenes, which can facilitate end-to-end stereo architecture learning and practical deployment.
\end{abstract}

}

\maketitle

\IEEEdisplaynontitleabstractindextext

%
\IEEEpeerreviewmaketitle

\IEEEraisesectionheading{\section{Introduction}\label{sec:introduction}}

%
%
%
%
\IEEEPARstart{D}{epth} serves as a realistic prerequisite for sensing the surrounding 3D scene structure in many high-level 3D vision tasks~\cite{chen20173d,saxena20083,gomez2019pl}. Image-based passive depth estimation approaches compare favorably to active sensors in terms of cost, working range, and flexibility. Among these methods, the well-posed stereo vision is preferentially chosen due to its straightforward settings, excellent accuracy, and reasonable cost.

The past decade has witnessed tremendous success in vision tasks with the help of deep learning, such as image classification~\cite{krizhevsky2012imagenet,szegedy2015going,he2016deep}, object detection~\cite{ren2016faster,he2015spatial}, and semantic segmentation~\cite{shelhamer2016fully,chen2017deeplab}. Benefiting from the convolutional neural networks (CNNs), deep stereo methods have also achieved encouraging progress~\cite{kendall2017end-to-end,chang2018pyramid,xu2020aanet,cheng2019learning,Zhang2019GANet,LEAstereo}, constantly improving challenging benchmarks like KITTI~\cite{kitti12,kitti15}. However, these methods suffer from performance degradation when deployed to unseen scenarios~\cite{tonioni2017unsupervised}, which is often caused by the gap between training and testing data domains, $e.g.,$ synthetic~\cite{sceneflow} and real-world data~\cite{kitti12,kitti15}. Domain-adaptive techniques~\cite{pang2018zoom,Liu2020StereoGAN,song2021adastereo} can perform well, but they are intrinsically dependent on the types of scenes available during training. Unfortunately, it is quite expensive and impractical to gather enough data from all available situations, such as different weather and road conditions in autonomous driving.

\begin{figure}[t]
\begin{center}
\includegraphics[height=5.64cm,width=8.2cm]{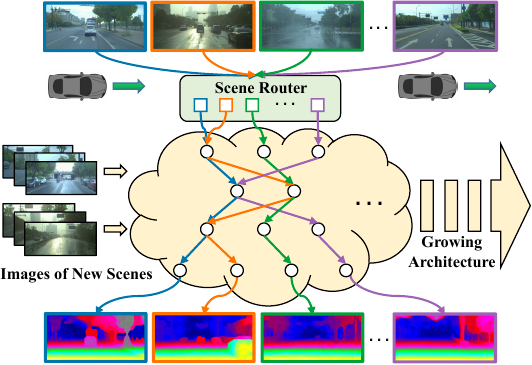}
\end{center}
   \caption{Schematic diagram of our framework deployed on real-world continuous driving scenes. The scene-specific architecture path chosen by Scene Router will be loaded for inference according to the scene type of input image.}
\label{obs_image}
\end{figure}

Imagine a car driving in real-world scenarios shown in Fig.~\ref{obs_image}. The car may go through continuous scenes changing from cloudy to rainy or from the city to the countryside. A stereo model with a single fixed architecture can hardly perform well in all types of scenes. Moreover, it is also difficult to continue learning new scenes without forgetting previously learned scenes. For optimal performance, an ideal model should grow its architecture as the number of scenes increases during training and adaptively load suitable architectures according to the scene type at deployment time. Additionally, when learning in new scenes, the model needs to avoid the performance drop in earlier scenes. 

Previous methods~\cite{MADNet,poggi2021continual} enable the stereo model to continuously adapt to current scenes by using an online learning scheme. Due to the fixed architecture, the model will inevitably forget previously learned scenes as new ones keep appearing. In contrast, we reformulate this problem as a continual stereo matching problem. By doing this, the model can \emph{continually} learn to estimate the disparity of new heterogeneous scenes while keeping in mind historical scenes with dynamic architectures. Without buffer time and online gradient updates, the model can swiftly adapt to rapidly changing \emph{continuous} scenes at inference. Furthermore, the model can generalize better to unseen scenes by leveraging previously learned scenes.

In this study, we propose a Reusable Architecture Growth (RAG) framework to address the continual stereo problem. RAG can overcome the catastrophic forgetting by freezing the model parameters learned in previous scenes. Since different scenes vary in color, illumination, and disparity distributions, we assign task-specific neural units for each new scene and adapt the model to them by architecture growth. To obtain a more compact architecture, we explicitly reuse the learned neural units during architecture growth, striking a balance between model performance and parameter efficiency. Since collecting ground truth labels requires expensive active sensors ($e.g.$, LIDAR) and manual intervention or post-processing~\cite{uhrig2017sparsity}, we further propose a proxy-supervision strategy to extend our framework to real-world label-free conditions. Under various challenging driving scenarios, our method achieves comparable or better performance than state-of-the-art methods. At deployment time, we propose a Scene Router module to automatically choose the scene-specific architecture path according to the scene type of the input.

Our contributions are summarized as follows:
\begin{itemize}
\setlength{\itemsep}{0pt}
\item We formulate the continual stereo matching problem that continually learns to estimate the disparity of new scenes without catastrophic forgetting.
\item A Reusable Architecture Growth framework consisting of task-specific neural unit search and architecture growth is introduced, which exhibits good reusability of previously learned units and is further extended to self-supervised conditions with a proposed proxy-supervision strategy.
\item A Scene Router module is designed to adaptively select the scene-specific architecture path for the current scene at inference, making the model quickly adapt to rapid scene switches and unseen scenes.
\item Comprehensive experiments across datasets demonstrate that our method is robust and scalable in different challenging driving scenarios.
\end{itemize}

This work extends our previous conference version~\cite{csm_zhang} in CVPR 2022 Oral methodologically and experimentally. For the methodological aspect, beyond the supervised continual stereo~\cite{csm_zhang}, we further formulate the self-supervised continual stereo problem and introduce a proxy-supervised architecture growth strategy by leveraging transferred synthetic driving data as a substitute for real-world data. In doing so, we can formulate a unified continual stereo problem for the completeness and practicality of the study. As for the experimental aspect, extensive experiments are conducted to demonstrate the effectiveness of our method. First, we add more real-world datasets to evaluate the proposed method under cross-scene and cross-dataset settings. Second, more visualization results are presented, including the searched structures of neural units and comparisons of qualitative disparity maps. Third, we discuss the impact of different scene learning orders on continual stereo performance, including easy-hard, repetition, and permutation orders, as well as model growth trends on various datasets. Fourth, ablation studies for self-supervised continual stereo are also reported. Fifth, the proposed framework is extended to monocular depth estimation and stereo-based 3D object detection tasks.

The remaining part of the paper is organized as follows. We review the related works in Section~\ref{related}. Then, we formulate the continual stereo problem in Section~\ref{continual}, followed by the developed method in Section~\ref{method}. Experimental results are conducted and analyzed in Section~\ref{experiment}, followed by discussion in Section~\ref{discussion} and extension to other tasks in Section~\ref{extention}. Finally, this paper is concluded in Section~\ref{conclusion}.

\section{Related Work}
\label{related}
\subsection{Deep Stereo Matching}
The deep neural network is first introduced in stereo only to calculate matching cost between patches~\cite{zbontar2015computing}, which relies on the traditional post-processing pipeline SGM~\cite{hirschmuller2007stereo} for refinement. DispNet~\cite{sceneflow} is the first end-to-end deep stereo model utilizing a correlation layer to encode matching information. Along this line, residual learning is exploited~\cite{pang2017cascade,liang2018learning} to obtain a more accurate disparity. Besides, semantic cues~\cite{yang2018SegStereo} and edge information~\cite{song2020edgestereo} are also incorporated. AANet~\cite{xu2020aanet} further boosts the performance through adaptive aggregation. 

Another category of deep stereo methods represented by GC-Net~\cite{kendall2017end-to-end} focuses on building a 4D cost volume and leveraging 3D convolutions to regress disparity. Subsequent improvements in the network structure include spatial pyramid feature extraction~\cite{chang2018pyramid}, convolutional spatial propagation~\cite{cheng2019learning}, and guided aggregation~\cite{Zhang2019GANet}. Other methods~\cite{guo2019group-wise,2020Cascade,Shen_2021_CVPR} explore to construct better cost volume, achieving remarkable performance on the KITTI benchmarks~\cite{kitti12,kitti15}. LEAStereo~\cite{LEAstereo} further improves the benchmark score through hierarchical neural architecture search. 

In addition to the pursuit of accuracy, there are also studies exploring lightweight or real-time stereo matching networks for better deployment in real environments~\cite{khamis2018stereonet,duggal2019deeppruner,tankovich2021hitnet} or improving the model generalizability from a sequence perspective~\cite{li2021revisiting}.

\begin{figure*}[t]
\begin{center}
\includegraphics[height=3.77cm,width=15cm]{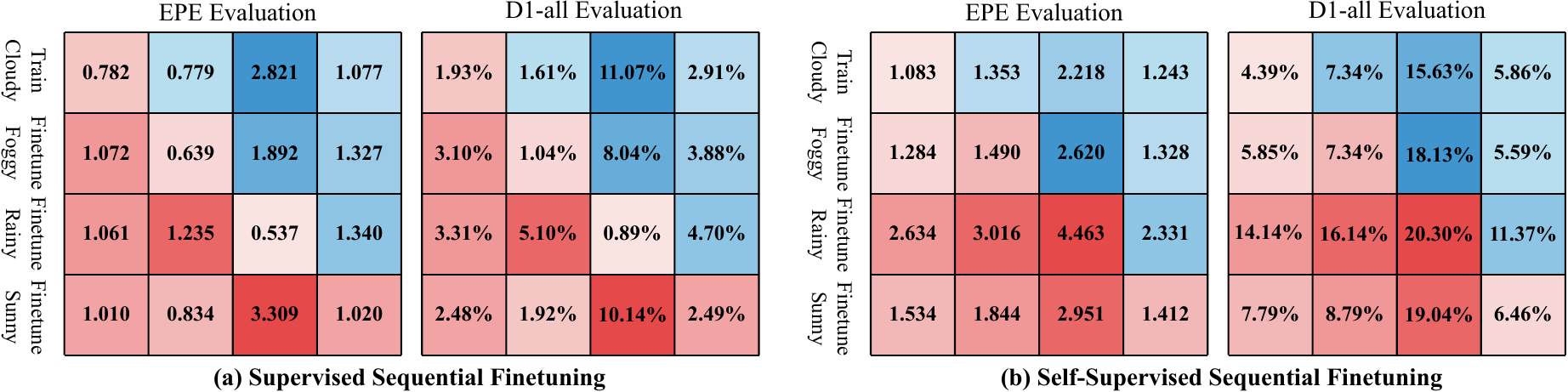}
\end{center}
   \caption{Catastrophic forgetting in stereo matching. The deep stereo model is first trained on the \emph{cloudy} scene and then finetuned on \emph{foggy}, \emph{rainy} and \emph{sunny} scenes in sequence. The red boxes refer to the performance on each scene learned so far, while the blue boxes refer to the generalization performance on unseen scenes. Light colors represent low errors.}
\label{obs_plot}
\end{figure*}

\subsection{Self-supervised and Adaptive Stereo}
Deep stereo models have achieved good performance benefiting much from the supervision of ground truth disparity. However, acquiring accurate disparity often requires expensive equipment and labor costs. Although synthetic datasets~\cite{sceneflow} provide large-scale image-disparity pairs for model pre-training, the pre-trained models will suffer from performance degradation when exposed to real-world scenarios~\cite{tonioni2017unsupervised}. To alleviate the problem, a viable way is to utilize photometric reconstruction through warping operations and disparity smoothing constraints~\cite{zhou2017unsupervised,godard2017unsupervised} to adapt to real-world scenarios in a self-supervised manner.

Another mainstream way is to leverage unsupervised domain adaptation methods to narrow the domain gap. Many efforts have been made to align the synthetic domain to the real-world domain at the input pixel level \cite{pang2018zoom,Liu2020StereoGAN} or internal feature space~\cite{song2021adastereo}. Apart from the offline learning mechanism, online learning can also be employed by temporal information exploitation~\cite{zhong2018open}, meta-learning scheme~\cite{tonioni2019learning}, or modular adaptation~\cite{MADNet,poggi2021continual}. However, neither self-supervised nor domain adaptive methods can fully overcome forgetting previously learned scenes and suffer from accumulation errors over time. In contrast, our method for continual stereo can learn from continuously collected data of new scenes without any forgetting while performing continuous inference and generalizing to unseen scenes based on previously learned scenes.

\subsection{Neural Architecture Search}
Neural architecture search (NAS) has attracted increasing interest in recent years. Most of the methods targeted on image classification tasks to search top-performing architectures by reinforcement learning~\cite{zoph2018learning,pham2018efficient}, evolutionary algorithms~\cite{real2019regularized}, and one-shot search~\cite{liu2018darts,zhang2020you}. In addition, some work also develop NAS to other tasks like semantic segmentation~\cite{liu2019auto,sun2021real} and object detection~\cite{chen2019detnas,wang2021fcos}. AutoDispNet~\cite{saikia2019autodispnet} first applies NAS to stereo matching by searching cell units of a U-shape architecture. Subsequently, LEAStereo~\cite{LEAstereo} achieves the top performance on several stereo benchmarks through hierarchical NAS. In this study, we leverage NAS to perform task-specific neural unit search and architecture growth for continually incremental scenes.

\subsection{Continual Learning}
Continual learning methods aim to overcome catastrophic forgetting, which are mainly divided into three families~\cite{delange2021continual}, $i.e.$, rehearsal, regularization, and parameter isolation. The rehearsal methods store representative samples in memory~\cite{rebuffi2017icarl} or construct pseudo-samples by generative models~\cite{shin2017continual}. The regularization-based methods~\cite{kirkpatrick2017overcoming,li2017learning,lopez2017gradient} preserve the previous knowledge by adding a regularization term to the loss function without extra memory requirements. Nevertheless, these two families still can not fully maintain the previously learned knowledge. 

To avoid any possible forgetting, the parameter isolation methods protect the model parameters of learned tasks when meeting new tasks. One way is to select a sub-network from a fixed super-network using a mask for each task~\cite{mallya2018packnet,fernando2017pathnet,serra2018overcoming}, but the fixed network has a limited capacity for continually increasing tasks. Another way utilizes dynamic architectures~\cite{rusu2016progressive,aljundi2017expert} by allocating a single model for each task separately, which does not consider the reusability of previous models. Other methods~\cite{li2019learn,wang2020lifelong} selectively expand new units or adapt from old units when learning a new task, which reduces the speed of model expansion to some extent. However, these methods all focus on image classification tasks. In this study, we advance a further step to the dense regression task for continual stereo matching and realize it with the high reusability of neural units.

\section{Continual Stereo Matching}
\label{continual}
\subsection{Catastrophic Forgetting in Stereo Matching}
We first demonstrate the catastrophic forgetting phenomenon in stereo matching, that is, deep stereo models with fixed architecture and parameter space often suffer from serious performance drop on previously learned scenes when adapting to new scenes.

For illustration, we construct a task sequence consisting of four kinds of weather conditions of the DrivingStereo~\cite{yang2019drivingstereo} dataset, $i.e.,$ \emph{cloudy} $\rightarrow$ \emph{foggy} $\rightarrow$ \emph{rainy}$ \rightarrow$ \emph{sunny}. The training and validation sets are divided for each scene. The off-the-shelf state-of-the-art model LEAStereo~\cite{LEAstereo} is chosen for our discussions. We first train the model on the \emph{cloudy} scene and then sequentially finetune it on the other scenes in both supervised and self-supervised manners. The evaluation is performed on each scene learned so far and unseen scenes, as shown in Fig.~\ref{obs_plot}. 

\begin{figure*}[t]
\begin{center}
\includegraphics[height=5.97cm,width=17cm]{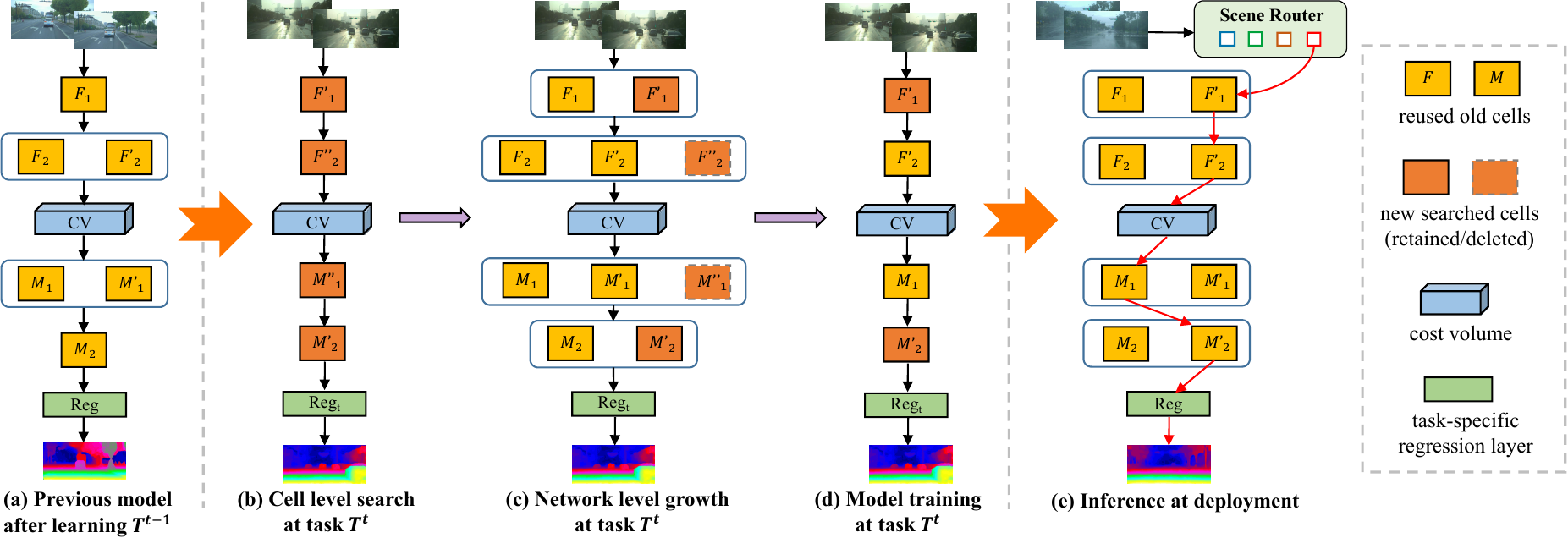}
\caption{Overview of our Reusable Architecture Growth framework. For the current task $\mathcal{T}^t$, based on the previous model (a), we first search task-specific neural units of the Feature Net (marked as $F$) and Matching Net (marked as $M$) (b), then select suitable units to make the network grow (c), and finally train the selected specific model (d). At test time, the scene-specific architecture path (marked in red) is selected for inference according to the Scene Router (e). Best viewed in color.}
\label{overview}
\end{center}
\end{figure*}

For supervised learning, we observe that sequential finetuning usually achieves the best performance after finetuning on the current scene, but it performs poorly on previously learned scenes or unseen scenes. This is especially obvious after finetuning on the \emph{rainy} scene since it is entirely different from the other three scenes in terms of color, illumination, and texture. As for self-supervised learning, better generalization performance is achieved in similar scenes like \emph{cloudy} and \emph{sunny} scenes. Unfortunately, when adapting to complex scenes like \emph{rainy} days, the model suffers from error accumulation and forgetting due to the lack of ground truth disparity, which results in performance deterioration over time. The experimental results show that catastrophic forgetting frequently occurs for both supervised and self-supervised stereo in heterogeneous driving scenes.

\subsection{Problem Statement}
The continual stereo problem $\mathcal{T}$ can be formulated as learning $N$ consecutive task sequences $\{\mathcal{T}^1, \mathcal{T}^2, ..., \mathcal{T}^N\}$ with each corresponding to a heterogeneous scene. Let $\Omega = \{\Omega^1, \Omega^2, ..., \Omega^N\}$ denote the data of the $N$ continual scenes, which is divided into training data $\Omega_{(train)}$ and testing data $\Omega_{(test)}$. For each task $\mathcal{T}^t$, the corresponding training data consist of $m^t$ triples, that is $\Omega^t_{(train)} = \{(I_{l_1}^t, I_{r_1}^t, D_{gt_1}^t), ..., (I_{l_{m^t}}^t, I_{r_{m^t}}^t, D_{gt_{m^t}}^t)\}$, where $I_l$, $I_r$, and $D_{gt}$ are left image, right image, and the ground truth disparity, respectively. The problem can be divided into supervised continual stereo and self-supervised continual stereo according to whether $D_{gt}$ is available during training.

Denote the learning model $h^t$ that contains all the parameters learned up to the current task $\mathcal{T}^t$. The model learns on the tasks from $\mathcal{T}^1$ to $\mathcal{T}^N$ sequentially, and only the training data of the current task $\Omega_{(train)}^t$ can be used. Both $\Omega_{(train)}^t$ and $\Omega_{(test)}^t$ will not be available in the subsequent tasks. The goal of the continual stereo is to continually learn the ability to estimate the disparity of incremental new scenes by \emph{maximizing} the performance of $h^t$ on the current task $\mathcal{T}^t$ while \emph{minimizing} the forgetting for previously learned tasks from $\mathcal{T}^1$ to $\mathcal{T}^{t-1}$. To achieve the maximum goal, we can minimize the following objective function,
\begin{equation}
\label{eq:global}
\mathcal{L}(h^{N} ; \Omega_{(train)})=\sum\nolimits_{t=1}^{N}\mathcal{L}_t \left(h^{t};\Omega_{(train)}^{t}\right).
\end{equation}

For supervised continual stereo, the objective can be specifically formulated as
\begin{equation}
\label{eq:sup}
\mathcal{L}_t\left(h^{t} ; \Omega_{(train)}^{t}\right)=\frac{1}{m^t} \sum\nolimits_{i=1}^{m^t} \mathcal{L}_{reg}\left(D_{pred_i}^t, D_{gt_i}^{t}\right),
\end{equation}
where $\mathcal{L}_{reg}$ is the smooth $l_1$ loss of the predicted disparity $D_{pred_i}^{t} = h^{t}\left(I_{l_i}^{t}, I_{r_i}^{t}\right)$ and the ground truth disparity $D_{gt_i}^{t}$. 
As for self-supervised continual stereo, the objective is formulated as
\begin{equation}
\label{eq:uns}
\mathcal{L}_t\left(h^{t} ; \Omega_{(train)}^{t}\right)=\frac{1}{m^t} \sum\nolimits_{i=1}^{m^t} \mathcal{L}_{self}\left(I_{l_i}^{t}, I_{r_i}^{t}, D_{pred_i}^t\right),
\end{equation}
where $\mathcal{L}_{self}$ consists of photometric reconstruction loss and disparity smoothness loss introduced in~\cite{godard2017unsupervised}.

Note that we have no access to all the training data at once, thus  Eq.~\eqref{eq:global} can not be directly optimized. Actually, for the current task $\mathcal{T}^t$, the parameters learned on the previous tasks have been optimized. By freezing the previously learned parameters, we can get an approximate optimization target of optimizing Eq.~\eqref{eq:sup} or Eq.~\eqref{eq:uns} for the current model $h^t$ while achieving the minimization goal.

\section{Method}
\label{method}
\subsection{Overview}
In this study, we propose the Reusable Architecture Growth to better deal with the current task.
Fig.~\ref{overview} shows the overall pipeline of our RAG. For the current task $\mathcal{T}^t$, we first search for task-specific neural units since different scenes have different colors, textures, and disparity distributions. To adapt the model to $\mathcal{T}^t$ without forgetting previous tasks, we then perform architecture growth to determine whether to reuse old units or use the searched new ones for each layer in the network level. The selected new units are retained while the unadopted are deleted. Finally, the model of the current task will be trained. At test time, we further introduce the Scene Router module to adaptively select a specific architecture path to predict the disparity of continuous image data stream. 

\begin{figure}[t]
\begin{center}
\includegraphics[height=4.61cm,width=8.5cm]{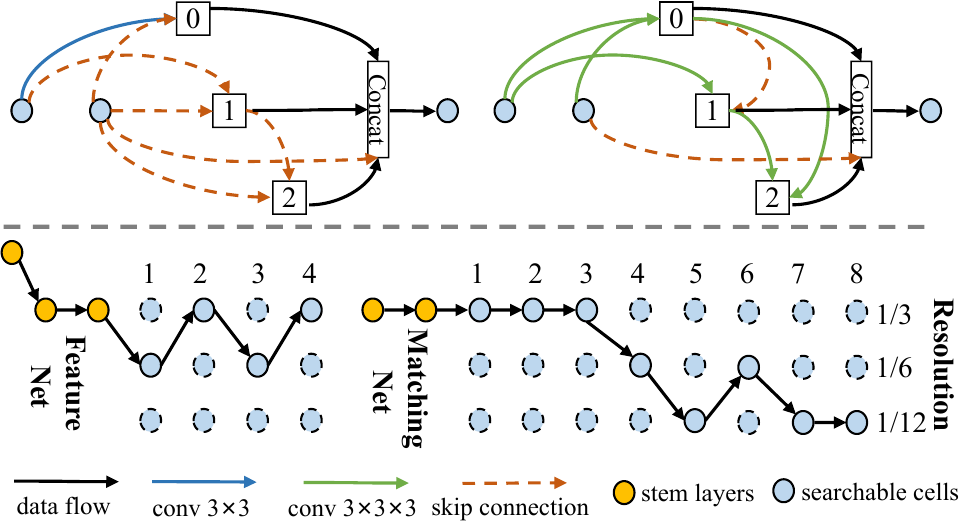}
\end{center}
   \caption{The network architecture of the base model.}
\label{base_model}
\end{figure}

\subsection{Basic Architecture}
Our base model comes from the variety of LEAStereo~\cite{LEAstereo} for its good scalability, including Feature Net, cost volume, Matching Net, and disparity regression layers. To better deploy on resource-limited edge devices, we adopt a lightweight version including 4-layer Feature Net and 8-layer Matching Net. Following the same training protocols, the searched architecture is shown in Fig~\ref{base_model}. The top two graphs are cell structures for Feature Net and Matching Net, respectively. The bottom is the whole network structure. The yellow dots refer to fixed "stem" structure, and the blue dots refer to searchable cells. There are three "stem" layers for the Feature Net, which are one $3 \times 3$ convolution layer with stride 3 and two $3 \times 3$ convolution layers with stride 1. For the Matching Net, the two "stem" layers are $3 \times 3 \times 3$ convolution layers with stride 1. In the continual stereo, the task-specific cell structures are searched for each scene.

\subsection{Reusable Architecture Growth}
\subsubsection{Cell Level Search}
The neural unit in our model is a searchable cell that consists of a fully-connected directed acyclic graph (DAG). Following~\cite{LEAstereo}, the cell consists of two input nodes from the preceding two layers, three intermediate nodes, and one output node. Let $\mathcal{O}$ denote a set of candidate operations. The set includes the $3 \times 3$ 2D convolution and skip connection for Feature Net. For Matching Net, it includes the $3 \times 3 \times 3$ 3D convolution and the skip connection. Then each intermediate node $s_j$ can be formulated as $s_{j}=\sum\nolimits_{i \sim j} o_{i j}\left(s_{i}\right)$, where $\sim$ indicates that node $i$ is connected to $j$, and $o_{ij}$ is the operation between them containing $K(=2)$ operations.

For each operation $o_{ij} = \{o_{ij}^1, ..., o_{ij}^K\}$, we allocate a validation score $\Delta=\{\delta_{ij}\}$, an iteration record $C = \{c_{ij}\}$, and a candidate probability $P = \{p_{ij}\}$. Both the validation score $\Delta$ and the iteration record $C$ are initialized to zero, and the candidate probability $P$ is uniformly initialized.

To update $P$, we apply the MdeNAS algorithm in \cite{zheng2019multinomial} to our cell level search with modifications. Define the selected set of operations after each sampling as $m^*$. Assuming the error rate (D1-all) on the validation set is $\sigma^*$ after one epoch, we have the validation score formulated as
\begin{equation}
\label{eq:simple}
\delta_{i j}^{m^*} = 1 - \sigma^*.
\end{equation}
The iteration record correspondingly increases by one. 

Intuitively, the operation with fewer iterations and higher validation scores should be preferentially selected, thus getting probability gain. Otherwise it should get probability decay for penalty. Following this point of view, the updating policy of $P$ is as follows:
\begin{equation}
\begin{aligned}
\label{eq:prob}
p_{ij}^{m^*} = p_{ij}^{m^*} + \alpha ( & \sum\nolimits_{k=1}^{K} \mathbb{I}(c_{i j}^{m^*}<c_{i j}^{k}, \delta_{i j}^{m^*}>\delta_{i j}^{k}) - \\
&\sum\nolimits_{k=1}^{K} \mathbb{I}(c_{i j}^{m^*}>c_{i j}^{k}, \delta_{i j}^{m^*}<\delta_{i j}^{k})),
\end{aligned}
\end{equation}
where $\mathbb{I(,)}$ denotes as the indicator function that equals to one if its condition is true and $\alpha$ is the update momentum (set to 0.01). To ensure that the sum of the probabilities is 1, we adopt the softmax operation on the candidate probability as $p_{i j}^{m^*}=\frac{\exp(p_{i j}^{m^*})}{\sum_{k=1}^{K} \exp(p_{i j}^{k})}$.
Finally, select the operations with the highest probability as the final operations of the cell. Extensive experiments demonstrate that task-specific neural units achieve superior performance to fixed units.

\subsubsection{Network Level Growth}
\label{sec:network_grow}
We define the network level search space as the arrangement of cells of old and new tasks. For the current task $\mathcal{T}^t$, the candidate architecture is composed of cells in each layer, $i.e.$, $h^t = \bigcup\nolimits_{j=1}^L \beta_j$, where $L$ is the number of layers and $\beta_j$ is composed of the reused old cells $\{o_j^1, ..., o_j^{t-1}\}$ and the searched new cell $o_j^t$. 

Similar to the cell level search, we also allocate a validation score $\Delta = \{\delta_j\}$, an iteration record $C = \{c_j\}$, and a selected probability $P = \{p_j\}$ to each cell. The update of the selected probability follows the policy in Eq.~\eqref{eq:prob}. Finally, the cells with the highest probability are selected to form the model $h^t$ of the current task $\mathcal{T}^t$.

\textbf{Growth Policy.} It is worth noting that the architecture growth will inevitably increase the model parameters. We hope to select reused old cells as many as possible to improve the reusability of the learned cells while maintaining good performance. To this end, we modify the updated policy in two aspects: initialization and validation score.

Unlike the cell level search, the reused cells learned on the old tasks have already been trained, so their iteration records should be initialized to a non-zero constant $c_0$ while the new cells are still initialized to zero. Besides, to improve the reuse rate of the old cells, the selected probability of old cells should be higher than that of new cells. Here, we initialize the probability of the old cells to $\gamma$ times that of the new cell, $i.e.$,
\begin{equation}
p_j^k =  \gamma \cdot p_j^t = \frac{\gamma}{\gamma(t-1)+1},  k \in \{1,2,...,t-1\}.
\end{equation}

As for the validation score, a simple way is to directly use the error rate like Eq.~\eqref{eq:simple}, but the reusability of the model is excluded. To keep the model more compact during growth, we explicitly integrate the model parameters into the growth evaluation. After each sampling, the selected architecture path is marked as $m^*$, and the number of parameters of the selected reused cells is $\phi^{m^*}$. Assuming that the error rate on the validation set after one epoch is $\sigma^*$. The validation score can be regarded as a function $f(\sigma^*, \phi^{m^*})$, which should obey three properties: i) $0 < f(\sigma^*, \phi^{m^*}) < 1$; ii) be negatively correlated with $\sigma^*$; iii) be positively correlated with $\phi^{m^*}$. Intuitively, we can utilize a simple weighted linear combination of $\sigma^*$ and $\phi^{m^*}$ to form the validation score as below:
\begin{equation}
\label{eq:alter}
\delta_j^{m^*} =  \mu \cdot \sqrt{(1- \sigma^*)} + (1 - \mu) \cdot \log(\frac{\phi^{m^*}}{\phi} + 1),
\end{equation}
where $\phi$ is the target number of parameters of the reused cells that controls the compactness of the model and $\mu$ is the weight (set to 0.9). However, this form requires hard trial for parameter tuning, which is inconvenient. To this end, we design the following validation score for better adaptability:
\begin{equation}
\label{eq:ours}
\delta_j^{m^*} = \sqrt{1- \sigma^*}\cdot \log(\frac{\phi^{m^*}}{\phi} + 1),
\end{equation}
The ablation study in Section~\ref{sec:ablation} shows that the above strategy improves the parameter efficiency while keeping satisfactory performance.

\subsection{Proxy-Supervised Architecture Growth}

In real-world driving scenarios, it is often difficult to obtain disparity labels for training since the LiDAR is bulky and expensive. To apply to the practical label-free environment, we move forward to extend our RAG framework to self-supervised continual stereo.

Specifically, we propose to transfer the scene style of synthetic driving images to be close to real-world images as the \emph{proxy supervision} for neural unit search and network growth. In general, the domain gap between synthetic and real-world data mainly lies in contents and colors. Fortunately, the synthetic datasets~\cite{sceneflow,cabon2020vkitti2,gaidon2016virtual} provide sufficient images of driving scenes that are consistent in content with real-world driving scenarios. To further narrow the color shift, we leverage the Progressive Color Transfer method~\cite{song2021adastereo} to convert synthetic images to scene-specific real-world images without adversarial training or extra parameters. By utilizing transferred synthetic driving images as the proxy supervision, our method can successfully search out scene-specific architectures that are better than the base model, as shown in Section~\ref{proxy}. Other operations during searching and growth keep the same. For the selected model of a specific scene, we first pre-train the model on transferred synthetic driving images and then adapt it to real-world images in a self-supervised manner.

Notably, previous self-supervised stereo methods pre-train the model on synthetic data and then adapt it to real-world data~\cite{zhou2017unsupervised,godard2017unsupervised}. They are feasible to be adapted to existing continual learning methods~\cite{kirkpatrick2017overcoming,rebuffi2017icarl,aljundi2017expert,li2019learn} by using the weights parameters pre-trained on synthetic data. However, it is impracticable for our framework since our method aims to search for cell-level structures suitable for each scene and the pre-trained parameters of fixed architectures are unusable. This further leads to another challenge: a randomly initialized stereo model can not be well trained from self-supervision via right-to-left image warping. In contrast, the proposed strategy alleviates both the unavailability of pre-trained parameters and training difficulties caused by random initialization.

\subsection{Scene Router}
At inference, a scene-specific architecture path needs to be selected to predict the disparity of continuous stereo images. Although the path can be manually chosen, it is often difficult for drivers to judge the current scene. To tackle the problem, we propose a Scene Router module to automatically select the path for the current scene.

As stated in~\cite{aljundi2017expert}, for the current task $\mathcal{T}^t$, we train a one-layer autoencoder for each task. The task-specific autoencoder takes the image feature representation $x_n^t$ as input and is trained using the mean square error loss $\mathcal{L}_{mse}$ between the reconstructed output $\hat{x}_n^t$ and the input $x_n^t$. At test time, the test image is input into different task-specific autoencoders to reconstruct itself, and the scene corresponding to the autoencoder with the smallest reconstruction error is selected as the final choice.

The above training mechanism can achieve good performance in image classification~\cite{aljundi2017expert}, but it fails in some cases of driving scenarios by experiments in Section~\ref{sec:ablation}. An autoencoder trained in one scene may yield a lower reconstruction error for the other due to the correlation between different scenes. To alleviate this problem, we advocate to explicitly enlarge the reconstruction loss of the image of the new scene on the previously trained autoencoders. Assuming that the reconstruction output obtained by the autoencoder trained on the old task $\mathcal{T}^i$ is $\hat{x}_o^i (i < t)$. Our goal is to make the reconstructed output $\hat{x}_n^t$ \emph{close} to the input $x_n^t$ and \emph{far} from $\hat{x}_o^i$, $i.e.,$ 
\begin{equation}
\operatorname{sim}_{nn}\left(\hat{x}_n^t, x_n^t \right) \gg \operatorname{sim}_{no}\left( \hat{x}_n^t, \hat{x}_{o}^{i}\right),
\end{equation}
where $\operatorname{sim}(x,y)$ is the similarity measurement using the inverse of the mean square error. Inspired by~\cite{oord2018representation}, we propose a \emph{scene contrastive loss} to impose this constraint as
\begin{small}
\begin{equation}
\mathcal{L}_{con} = - \log \left(\frac{\exp (\operatorname{sim}_{n n} / \tau )}{\exp (\operatorname{sim}_{n n} / \tau)+\sum_{i} \exp (\operatorname{sim}_{n o} / \tau)}\right),
\end{equation}
\end{small}
where $\tau$ is the temperature (typically set to 2). Thus the entire training loss becomes to $\mathcal{L}_{auto} = \mathcal{L}_{mse} + \lambda \cdot \mathcal{L}_{con}$, where $\lambda = 0.1$. Experiments in Section~\ref{sec:ablation} show that $\mathcal{L}_{con}$ can significantly improve the accuracy of scene division.

\section{Experiments}
\label{experiment}
\subsection{Datasets and Evaluation Metrics}
Here, we describe the datasets and evaluation metrics used for the experiments.

\textbf{DrivingStereo~\cite{yang2019drivingstereo}.} A collection of large-scale real-world outdoor stereo image pairs with sparse ground truth disparity maps. This dataset contains about 17K sequential training pairs and about 8K testing pairs. From the entire dataset, we use the specially selected 2000 frames with four kinds of weather (\emph{cloudy}, \emph{foggy}, \emph{rainy}, \emph{sunny}) for continual stereo. Each scene includes 500 stereo pairs with 400 pairs for training and 100 pairs for testing. The resolution of the image is $881 \times 400$.

\begin{table*}[!htbp]
\begin{center}
\caption{Results of cross-scene comparisons for supervised continual stereo on DrivingStereo, KITTI raw, and Virtual KITTI datasets. Red and blue represent the best and the second-best results on the corresponding datasets.}
\label{incremental}
\small
\setlength{\tabcolsep}{1.8mm}{
\begin{tabular}{ c  | c  c  c  c  | c  c  c  c | c c c c }
\toprule
{\multirow{3}{*}{Methods}} & \multicolumn{4}{ c | }{DrivingStereo} & \multicolumn{4}{ c|  }{KITTI raw} & \multicolumn{4}{ c  }{Virtual KITTI}\\
  &  \multicolumn{2}{ c }{FAE} & \multicolumn{2}{ c | }{BWT} & \multicolumn{2}{ c  }{FAE} & \multicolumn{2}{ c | }{BWT} & \multicolumn{2}{ c }{FAE}  & \multicolumn{2}{ c  }{BWT} \\
  & EPE$\downarrow$ & D1$\downarrow$ & EPE$\downarrow$ & D1$\downarrow$  & EPE$\downarrow$  & D1$\downarrow$ & EPE$\downarrow$ & D1$\downarrow$  & EPE$\downarrow$ & D1$\downarrow$ & EPE$\downarrow$  & D1$\downarrow$ \\
\midrule
Incremental Finetuning & 0.937 & 2.60\%  & 0.342 & 1.43\% & 0.594 & 0.78\% & 0.050 & 0.04\% & 0.892 & 5.63\% & 0.353 & 2.66\%  \\
EWC~\cite{kirkpatrick2017overcoming} & 1.053 & 4.02\% & 0.095 & 0.56\% & 0.620 & 0.90\% & 0.008 & 0.02\%  & 0.934 & 5.85\% & 0.070 & 0.45\%  \\
iCaRL~\cite{rebuffi2017icarl} & 0.866 & 1.78\% & 0.225 & 0.50\% & 0.586 & 0.72\% & 0.011 & 0.03\% & 0.863 & 5.01\% & 0.143 & 0.80\%    \\
Expert Gate~\cite{aljundi2017expert} & 0.681 & 1.52\% & 0.0 & 0.0\% & 0.586 & 0.74\% & 0.0 & 0.0\% & \textcolor{red}{0.697} & \textcolor{red}{3.91\%} &  0.0 & 0.0\% \\
Learn to Grow~\cite{li2019learn} & \textcolor{blue}{0.662} & \textcolor{blue}{1.34\%}  &  0.0 & 0.0\% & 0.598 & 0.77\% & 0.0 & 0.0\% & 0.836 & 5.03\% & 0.0 & 0.0\% \\
Joint Training (Ideal) & 0.673 & 1.38\% & - & -  &  \textcolor{red}{0.554} & \textcolor{red}{0.64\%} & - & -  & 0.737 & \textcolor{blue}{4.18\%} & - & - \\
\midrule
RAG (Ours) & \textcolor{red}{0.637} & \textcolor{red}{1.21\%} & 0.0 & 0.0\% &  \textcolor{blue}{0.583} &  \textcolor{blue}{0.71\%} & 0.0 & 0.0\% & \textcolor{blue}{0.717} & \textcolor{blue}{4.18\%} & 0.0 & 0.0\%  \\
\bottomrule
\end{tabular}}
\end{center}
\end{table*}

\textbf{KITTI Raw~\cite{geiger2013vision}.} A collection of real-world outdoor stereo video sequences covering heterogeneous environments, namely \emph{residential}, \emph{city}, \emph{road} and \emph{campus}. This dataset contains $\sim$ 43k video frames with sparse depth labels \cite{uhrig2017sparsity} converted into disparities by knowing the camera parameters. Since real-world labels are hard to obtain, we extract two videos from each scene for training ($\sim$ 1000 pairs) and testing ($\sim$ 100 pairs) to mimic the real environments for continual stereo. The resolution of the image is about $1248 \times 384$.

\textbf{Virtual KITTI~\cite{cabon2020vkitti2,gaidon2016virtual}.} A collection of the synthetic clone of the real KITTI dataset containing five sequences of Scene 01, 02, 06, 18, and 20. Each scene has nine variants with five different weather and four modified camera configurations. Since the weather conditions have been considered in DrivingStereo, we select the five scenes with four camera configurations and divide them into 80\% training sets and 20\% testing sets for supervised continual stereo.

\textbf{SceneFlow~\cite{sceneflow}.} A collection of large-scale synthetic images with accurate and dense disparity maps produced by rendering 3D models for training deep stereo models. This dataset contains 35454 training image pairs and 4370 testing image pairs with a resolution of $960 \times 540$, which are divided into three subsets, named \emph{FlyingThings3D}, \emph{Monkaa} and \emph{Driving}. For self-supervised continual stereo, we perform neural units search on the \emph{Driving} subsets and perform task-specific model pre-training on the whole training sets.

\textbf{CityScapes~\cite{cordts2016cityscapes}.} A collection of real-world outdoor street scene images in different cities, providing stereoview images with a resolution of $2048 \times 1024$ and precomputed depth maps using SGM~\cite{hirschmuller2007stereo}. From the whole training sets, we select four kinds of city scenes with most samples for self-supervised continual stereo, namely \emph{bremen}, \emph{dusseldorf}, \emph{hamburg} and \emph{strasbourg}. We split the data for each city scene into 80\% training sets and 20\% testing sets and align the image resolution with that of KITTI.

\textbf{KITTI 2012/2015~\cite{kitti12,kitti15}.} A collection of real-world outdoor driving scenes with two versions: the KITTI 2012 dataset provides 194 stereo training images with sparse ground truth disparity and 195 testing pairs for benchmark evaluation, and the KITTI 2015 dataset provides 200 training and 200 testing pairs. We combine the training sets of two versions with 90\% training sets and 10\% testing sets for cross-dataset continual stereo evaluation.

\textbf{Metrics.}
The stereo performance is accessed using the standard end-point-error (EPE) and D1-all metrics. The EPE metric measures the average pixel error between the predicted and ground truth disparity. As for the D1-all metric, we calculate the percentage of pixels with an absolute disparity error larger than 3 and a relative error larger than 5\%.
For continual stereo evaluation, we use Final Average Error (FAE) to represent the average performance of the final model on each task learned so far. Backward Transfer
(BWT) introduced in~\cite{lopez2017gradient} is also adopted to evaluate the forgetting of previously learned knowledge.
To evaluate the reusability of the learned cells, we further denote a metric named Average Reuse Rate (ARR) to calculate the average proportion of the parameters of the old cells in the current architecture.

\subsection{Evaluation Protocols and Baselines}
Our method is implemented with PyTorch~\footnote{Code is available at https://github.com/chzhang18/RAG.}. The training protocols follow our conference version~\cite{csm_zhang}. The evaluation is divided into two cases, namely, the cross-scene comparison of different scenes of the same dataset and the cross-dataset comparison between different scenes of different datasets. The learning sequence for cross-dataset evaluation is KITTI 2012/2015 $\rightarrow$ DrivingStereo (\emph{cloudy}) $\rightarrow$ KITTI raw (\emph{campus}) $\rightarrow$ CityScapes (\emph{bremen}). More training
details can be found in the supplementary material of the conference version~\cite{csm_zhang}.

We compare our method with several representative continual learning methods, which are regularization-based EWC~\cite{kirkpatrick2017overcoming}, rehearsal-based iCaRL~\cite{rebuffi2017icarl}, as well as Expert Gate~\cite{aljundi2017expert} and Learn to Grow~\cite{li2019learn} based on dynamic structure. We also adopt the other two widely used baselines, incremental finetuning and the joint training that uses all data from various scenes (ideal conditions). 

We adopt the base model to reimplement these baselines for continual stereo evaluation. For iCaRL, we set the size of the memory bank to 10\% of the training set of each task. For Expert Gate, we use the output disparity of the previous models on the new scene as the pseudo-label for distillation~\cite{li2017learning}. For Learn to Grow, we reimplement it using the same search strategy as in our cell-level search. Notably, for self-supervised continual stereo, all these methods are initialized with the weights parameters pre-trained on the synthetic dataset~\cite{sceneflow} for 20 epochs using the architecture in Fig.~\ref{base_model}. They are then trained in a self-supervised manner using photometric reconstruction and disparity smoothness constraints~\cite{zhou2017unsupervised,godard2017unsupervised}.

\begin{figure}[t]
\begin{center}
\includegraphics[height=12.49cm,width=8.5cm]{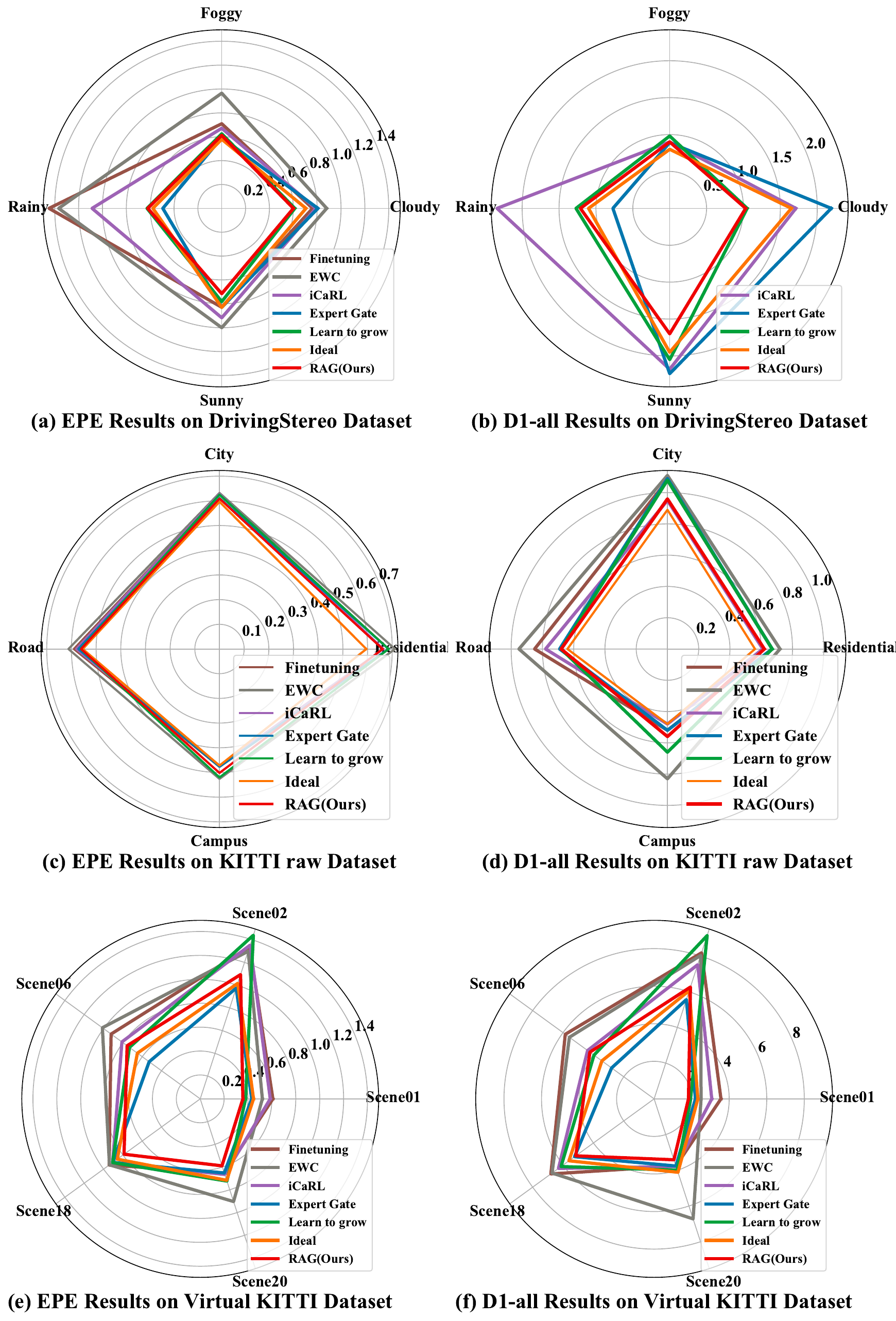}
\end{center}
   \caption{Cross-scene comparison results on each scene of DrivingStereo, KITTI raw, and Virtual KITTI datasets for supervised continual stereo.}
\label{continual_plot}
\end{figure}

\subsection{Supervised Continual Stereo Evaluation}

\subsubsection{Cross-Scene Comparisons}
The cross-scene comparison results with baselines are listed in Table~\ref{incremental}. It is clear that our method achieves remarkable performance on all three datasets across various weather and road conditions. Benefiting from the task-specific neural units, our RAG outperforms Learn to Grow and achieves the best performance on DrivingStereo. A surprising result is that our RAG even surpasses the performance of joint training in ideal on DrivingStereo and Virtual KITTI except for KITTI raw. This suggests that joint training on all the scenes may cause the model only to find the global solution of the mixed scenes, but the solution is not always optimal for each scene, especially when the difference across scenes is significant. On the contrary, learning each scene in sequence by architecture growth can decouple the mixed scenes to find a local solution for each scene. The claim is further validated that Expert Gate obtains superior results on Virtual KITTI since it allocates a separate base model to each scene. Each base model can better focus on the current scene under sufficient heterogeneous data, but the number of parameters will increase sharply simultaneously. By contrast, our method can keep high reusability during architecture growth while achieving competitive performance with fewer parameters. For catastrophic forgetting, finetuning suffers the most, as evidenced by the highest BWT. Compared with regularization and rehearsal based methods, our method does not forget any previous knowledge since our BWT is zero.

\begin{table}[!htbp]
\begin{center}
\caption{Results of cross-dataset comparisons for supervised continual stereo.}
\label{cross_dataset}
\small
\setlength{\tabcolsep}{1.8mm}{
\begin{tabular}{ c  | c  c | c  c   }
\toprule
{\multirow{2}{*}{Methods}}  &  \multicolumn{2}{ c| }{FAE} & \multicolumn{2}{ c  }{BWT} \\
  & EPE$\downarrow$ & D1$\downarrow$ & EPE$\downarrow$ & D1$\downarrow$  \\
\midrule
Incremental Finetuning & 3.064 & 10.43\% & 2.972 & 9.66\%  \\
EWC~\cite{kirkpatrick2017overcoming} & 1.590 & 7.83\% & 0.278 & 2.13\%  \\
iCaRL~\cite{rebuffi2017icarl} & 1.746 & 5.79\% & 1.214 & 3.47\%     \\
Expert Gate~\cite{aljundi2017expert} & 0.880 & 3.45\% & 0.0 & 0.0\%  \\
Learn to Grow~\cite{li2019learn} & \textcolor{blue}{0.865} & \textcolor{blue}{3.38\%} & 0.0 & 0.0\%    \\
Joint Training (Ideal)  & 0.962 & 3.71\% & - & - \\
\midrule
RAG (Ours) &  \textcolor{red}{0.846} & \textcolor{red}{3.12\%} & 0.0 & 0.0\%  \\
\bottomrule
\end{tabular}}
\end{center}
\end{table}

\begin{figure}[t]
\begin{center}
\includegraphics[height=3.02cm,width=8.5cm]{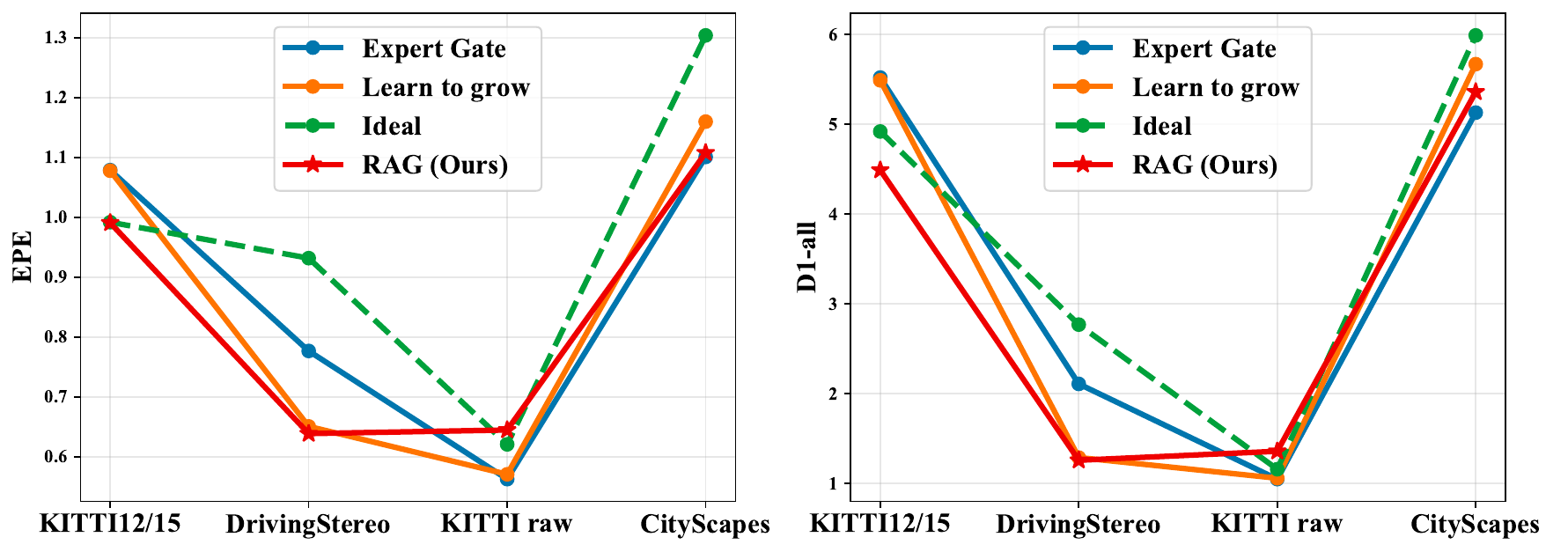}
\end{center}
   \caption{Comparison with comparable baselines on each scene of different datasets for supervised continual learning.}
\label{crossdata_sup_plot}
\end{figure}

We further compare the FAE on each scene with baselines as shown in Fig.~\ref{continual_plot}. The D1-all results of incremental finetuning and EWC are excluded on DrivingStereo due to poor performance. Our RAG achieves the best results in most scenes on all the datasets despite being inferior in some scenes, such as \emph{rainy} on DrivingStereo, \emph{campus} on KITTI raw, and Scene 06 on Virtual KITTI. It seems that the knowledge learned from previous scenes is not always sufficient to support the model to achieve the best in each scene under reusable constraints. Despite this, we still achieve the top performance regarding the final average error.

\subsubsection{Cross-dataset Comparisons}
Table~\ref{cross_dataset} shows the cross-dataset comparison results with various baselines. Our proposed RAG outperforms all the compared methods on all the metrics. We observe that regularization-based (EWC) and rehearsal-based methods (iCaRL) obtain high BWT and poor FAE. The large gap between different scenes from various datasets makes these methods suffer from severe forgetting. An interesting result appears again that dynamic architecture based methods outperform joint training in ideal. This verifies our previous analysis that the global solution for heterogeneous scenes may not be optimal for each scene. By contrast, learning each scene in order allows the model to focus on the current scene better.

We further demonstrate the performance comparisons of each scene with several comparable methods, as shown in Fig.~\ref{crossdata_sup_plot}. Our method performs best on three of the four datasets except KITTI raw. RAG searches for suitable cell-level architectures for each scene compared to Learn to Grow while achieving better knowledge transfer by leveraging old neural units learned in previous scenes compared to Expert Gate. These two mechanisms enable our method to achieve optimal performance.

\begin{table*}[!htbp]
\begin{center}
\caption{Ablation study of neural unit search and architecture growth on the DrivingStereo dataset.}
\label{ablation}
\small
\setlength{\tabcolsep}{1.8mm}{
\begin{tabular}{  c c c  | c  | c  c  |   c  c  | c  c  |  c c  |  c c }
\toprule
Cells & Fea. Net & Mat. Net & \# param & \multicolumn{2}{ c | }{Cloudy} &  \multicolumn{2}{ c|  }{Foggy} &  \multicolumn{2}{ c| }{Rainy}  &   \multicolumn{2}{ c | }{Sunny}  &  \multicolumn{2}{ c  }{FAE}  \\
Search & Grow & Grow  & (M) & EPE $\downarrow$ & D1 $\downarrow$ & EPE $\downarrow$ & D1 $\downarrow$  & EPE $\downarrow$ & D1 $\downarrow$  & EPE $\downarrow$ & D1 $\downarrow$ & EPE $\downarrow$ & D1 $\downarrow$\\
\midrule
 &  &  & 0.61 & 0.782 & 1.93\%  & 0.639 & 1.03\% & 0.541 & 0.93\% & 1.020 & 2.49\%  & 0.746 & 1.60\%  \\
\Checkmark &  &  & 0.61 & 0.601 & 1.03\% & 0.603 & 0.82\% & 0.546 & 0.99\% & 0.650 & 1.34\%  & \textbf{0.600} & \textbf{1.05\%}\\
\midrule
\Checkmark\ & \Checkmark &  & 0.17 & 0.733 & 1.74\% & 0.704 & 1.40\% & 0.806 & 2.30\% & 0.696 & 1.66\% & 0.735 & 1.78\% \\
\Checkmark &  & \Checkmark & 0.38 & 0.685 & 1.40\% & 0.712 & 1.39\% & 0.585 & 1.08\% & 0.707 & 1.69\% & 0.672 & 1.39\% \\
 & \Checkmark & \Checkmark & 0.41 & 0.618 & 1.03\%  & 0.604  & 0.84\% & 0.645 & 1.33\% & 0.753 & 1.90\% & 0.655 & 1.28\% \\
\Checkmark & \Checkmark & \Checkmark & 0.48 & 0.601 & 1.03\% & 0.612 & 0.90\% & 0.620 & 1.21\% & 0.715 & 1.70\%  & \textbf{0.637} & \textbf{1.21\%}\\
\bottomrule
\end{tabular}}
\end{center}
\end{table*}

\subsubsection{Ablation Study}
\label{sec:ablation}

\textbf{Neural Unit Search.}
In continual stereo, data from heterogeneous scenes are quite different in color, illumination, and disparity distribution. Hence, it is better to design task-specific neural units for each scene. For illustration, we train a separate model with fixed cells (searched on synthetic data) and searchable cells for each scene. The experimental results are listed in the top two rows of Table~\ref{ablation}. It is clear that using task-specific cells achieves remarkable advantages in three of four scenes. Even in the \emph{rainy} scene, we achieve considerable performance. This indicates that searching for task-specific cells is better than using fixed cells. 

\begin{figure}[t]
\begin{center}
\includegraphics[height=8.18cm,width=8cm]{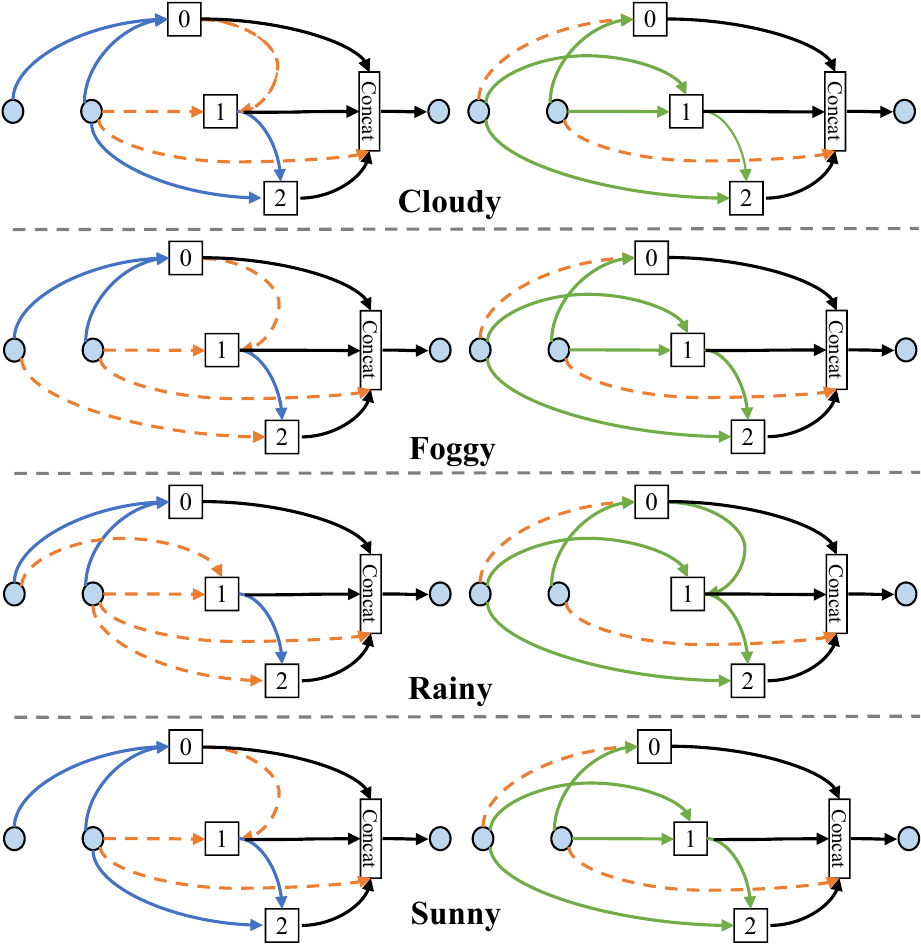}
\end{center}
   \caption{Visualization of searched cell structures on four scenes of the DrivingStereo dataset. The left column is for the Feature Net, and the right is for the Matching Net.}
\label{vis_cell}
\end{figure}

\begin{table}[t]
\caption{Ablation results of different strategies for the model reusability of supervised continual stereo on the DrivingStereo dataset.}
\small
\subfloat[Comparison results using different initialization values. \label{tab_supa}]{
\setlength{\tabcolsep}{4.5mm}{
\begin{tabular}{l | c c c }
\toprule
Initial Value & EPE $\downarrow$ & D1 $\downarrow$ & ARR $\uparrow$\\
\midrule
$c_0 = 0$ &  0.818 & 2.12\% &  \textbf{81.5\%} \\
$c_0 = 10$ & \textbf{0.655} & \textbf{1.34\%} & 40.7\% \\
$c_0 = 20$ & 0.676 & 1.42\% & 31.5\% \\
\bottomrule
\end{tabular}}}\hspace{3mm}
\subfloat[Comparison results using different probability multiples. \label{tab_supb}]{
\setlength{\tabcolsep}{5mm}{
\begin{tabular}{l | c c c}
\toprule
Multiple & EPE $\downarrow$ & D1 $\downarrow$ & ARR $\uparrow$\\
\midrule
$\gamma = 1$ &  \textbf{0.655} &  \textbf{1.34\%} &  40.7\% \\
$\gamma = 2$ &  0.671 & 1.36\% &  48.1\% \\
$\gamma = 5$ &  0.703 & 1.55\% &  49.3\% \\
\bottomrule
\end{tabular}}}\hspace{3mm}
\subfloat[Comparison results using different validation scores. \label{tab_supc}]{
\setlength{\tabcolsep}{3.5mm}{
\begin{tabular}{c | c c c}
\toprule
Validation Score & EPE $\downarrow$ & D1 $\downarrow$ & ARR $\uparrow$\\
\midrule
Eq.~\eqref{eq:simple} & 0.671 &  1.36\% &  48.1\% \\
Eq.~\eqref{eq:alter} & 0.731 &  1.78\% &  \textbf{54.1\%} \\
Eq.~\eqref{eq:ours} & \textbf{0.660}  &  \textbf{1.32\%}  &  50.9\% \\
\bottomrule
\end{tabular}}}\hspace{3mm}
\subfloat[Comparison results using different target parameters. \label{tab_supd}]{
\setlength{\tabcolsep}{3.5mm}{
\begin{tabular}{c | c c c}
\toprule
Target Parameters & EPE $\downarrow$ & D1 $\downarrow$ & ARR $\uparrow$\\
\midrule
$\phi = \Phi/3$ & 0.664 & 1.40\% & 48.2\% \\
$\phi = \Phi/2$ & 0.637 & 1.21\% &  \textbf{50.1\%} \\
$\phi = \Phi$ & \textbf{0.660}  &  \textbf{1.32\%}  &  50.9\% \\
\bottomrule
\end{tabular}}}\hspace{3mm}
\label{tab_sup}
\end{table}

Fig.~\ref{vis_cell} presents the visual structure of neural units searched in each scene. We are surprised to find that the cell structures of the \emph{cloudy} and \emph{sunny} scenes are the same, which indicates that the two scenes are highly similar. The Feature Net of the \emph{foggy} scene is slightly different from the previous two scenes, but they share the same Matching Net. The unique scene is the \emph{rainy} scene, whose cell structures of both networks are different from the other three scenes. This suggests that the rainy scene presents completely different data distributions from other scenes. Notably, these cell structures searched on real-world data are different from those on synthetic data in Fig.~\ref{base_model}. This further shows that cell structures correlate with scene types highly. For our RAG framework, we search for task-specific cell structures for each scene, thus achieving better performance.

\textbf{Architecture Growth.}
The architecture to grow comes from Feature Net and Matching Net. The results in the last four rows in Table~\ref{ablation} demonstrate the contributions of each part. Only the Feature Net growth can achieve reasonable performance, but it is inferior to the model using fixed cells. Only the Matching Net growth achieves better performance benefiting from the more powerful 3D convolutions. The growth of both parts further reduces the error rates and EPE. This suggests that the simultaneous growth of the two parts can maximize the model performance on the continual stereo. Our RAG even achieves comparable performance to the multiple models with task-specific cells ($2^{nd}$ row) yet has fewer parameters. 

\begin{figure}[t]
\begin{center}
\includegraphics[height=2.78cm,width=8.5cm]{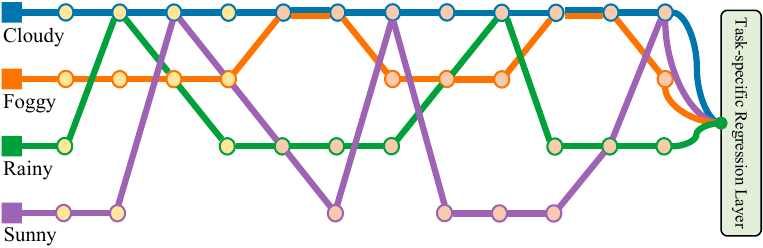}
\end{center}
   \caption{Visualization of the growing architecture on the DrivingStereo dataset. The architecture paths in four colors represent the inference paths of four kinds of weather. We only depict cells in the Feature Net (yellow circle) and Matching Net (red circle).}
\label{vis-reuse}
\end{figure}

\textbf{Reusability.}
Our method will inevitably increase the model parameters as the scenes increase. To tackle this problem, we propose a series of strategies to improve the model reusability while maintaining good performance. Here, four factors are considered, $i.e.$, the initial value $c_0$, the multiple $\gamma$, the validation score $\delta_j^{m^*}$, and the target parameters $\phi$.

Table~\ref{tab_supa} shows the results of several different initial values of $c_0$. Zero initialization for the well-trained old cells can get a high ARR, but the model's performance drops rapidly as the scenes increase. In contrast, non-zero initialization achieves more stable performance despite decreasing the ARR. We set $c_0 = 10$ in this work. 

To improve the reusability of the old cells, we set their initial probability to $\gamma$ times that of the new cells. One can observe in Table~\ref{tab_supb} that larger multiples can increase the ARR, but the error rates also increase. We choose $\gamma = 2$ as the trade-off between performance and reuse rates.

Table~\ref{tab_supc} compares the results of different validation scores. Compared with using only the error rate like Eq.~\eqref{eq:simple}, our proposed design in Eq.~\eqref{eq:ours} improves both the performance and ARR. Our strategy also achieves better performance than linear weighting in Eq.~\eqref{eq:alter} despite slightly less ARR. This indicates that our design can better select suitable old cells for the new task during architecture growth.

\begin{figure}[t]
\begin{center}
\includegraphics[height=3.41cm,width=8.6cm]{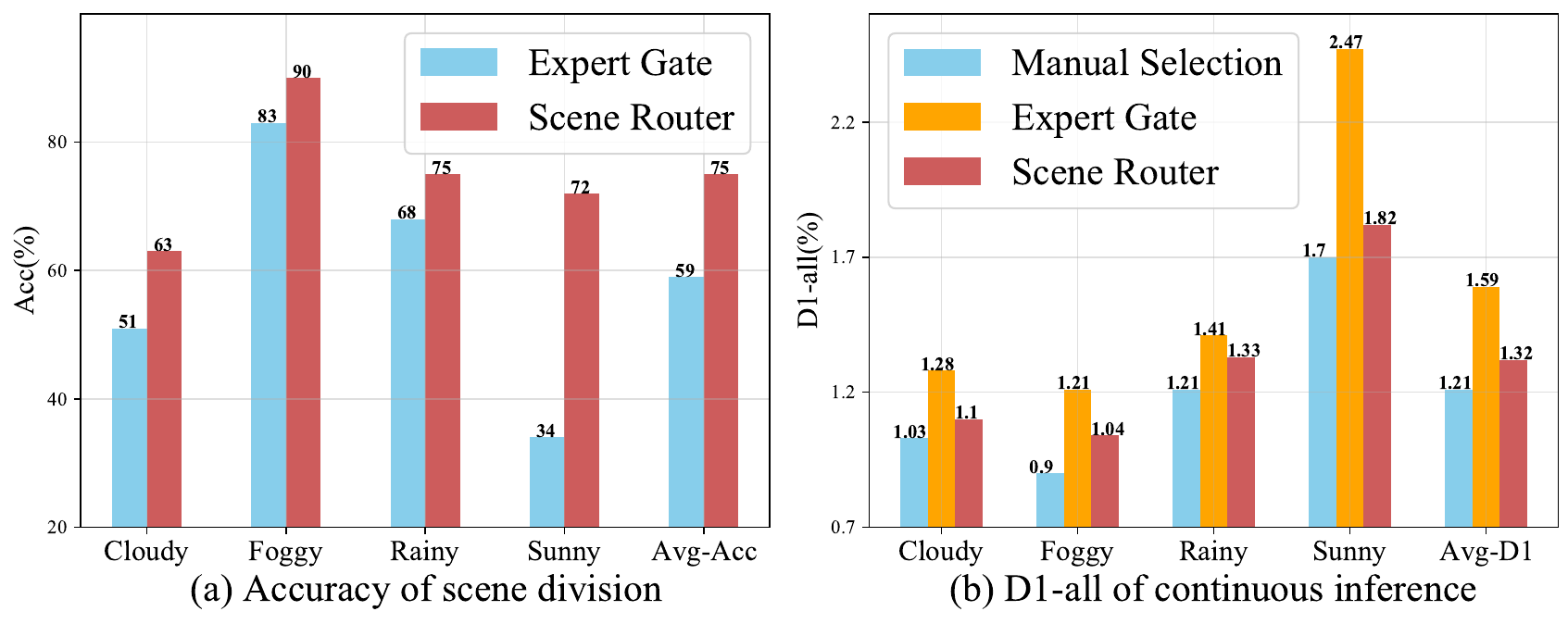}
\end{center}
   \caption{The effectiveness of Scene Router for scene division and continuous inference on the DrivingStereo dataset.}
\label{router-router}
\end{figure}

\begin{table}[!htbp]
\begin{center}
\caption{Cross-scene comparison results with the Mixture of Experts model on the DrivingStereo dataset.}
\label{moe}
 \normalsize
\setlength{\tabcolsep}{4mm}{
\begin{tabular}{ c  | c  c   }
\toprule
Methods & EPE$\downarrow$ & D1$\downarrow$ \\
\midrule
Expert Gate~\cite{aljundi2017expert} & 0.681 & 1.52\% \\
Joint Training (Ideal) & 0.673 & 1.38\%  \\
Mixture of Experts (MoE) & \textcolor{blue}{0.652} & \textcolor{red}{1.29\%}  \\
\midrule
RAG &  \textcolor{red}{0.648}  &  \textcolor{blue}{1.32\%} \\
\bottomrule
\end{tabular}}
\end{center}
\end{table}

We further explore the impact of the target number of parameters of the old cells $\phi$. As listed in Table~\ref{tab_supd}, using a larger target number can get a higher reuse rate, but the error rate does not decrease linearly. We finally choose the suitable intermediate value $\phi = \Phi/2$ where $\Phi$ represents the number of parameters of a base model. 

The visualization of the growing architecture in Fig.~\ref{vis-reuse} gives a more intuitive description of the reusability of old cells. Initially, the model builds a separate architecture path on the \emph{cloudy} scene. Starting from the second scene, the model begins to reuse old cells from the previous scenes. For example, the green path shares nodes with the blue and orange paths, which indicates that the model reuses the cells learned in the \emph{cloudy} and \emph{foggy} scenes when learning the \emph{rainy} scene. Overall, the average reuse rate of the model reaches 50\% after learning four scenes, which benefits from the proposed reusable architecture growth strategy.

\textbf{Scene Router.}
At deployment, we adopt Scene Router to adaptively select the scene-specific architecture path for continuous disparity estimation. Fig.~\ref{router-router}(a) shows the accuracy of scene division of our method and Expert Gate~\cite{aljundi2017expert}. Serious misjudgments occur in Expert Gate in a certain scene like \emph{sunny}, which makes the model corresponding to sunny days unable to be utilized well. In contrast, our Scene Router can achieve balanced results in various scenes and improve the mean accuracy by about 15\% benefiting from the proposed scene contrastive loss. Despite our method failing in some cases, it does not mean that the final stereo performance will drop significantly. Actually, images in different scenes are sometimes difficult to distinguish clearly, such as \emph{sunny} and \emph{cloudy} scenes. The misclassification of these two scenes may yield comparable stereo performance. To this end, we further compare our Scene Router with Expert Gate and manual selection. As depicted in Fig.~\ref{router-router}(b), Scene Router significantly surpasses Expert Gate, achieving considerable performance compared to manual selection.

\begin{table*}[!htbp]
\begin{center}
\caption{Results of cross-scene comparisons for self-supervised continual stereo on DrivingStereo, KITTI raw, and CityScapes datasets. Red and blue represent the best and the second-best results on the corresponding datasets.}
\label{self-incremental}
\small
\setlength{\tabcolsep}{1.8mm}{
\begin{tabular}{ c  | c  c  c  c  | c  c  c  c | c c c c }
\toprule
{\multirow{3}{*}{Methods}} & \multicolumn{4}{ c | }{DrivingStereo} & \multicolumn{4}{ c|  }{KITTI raw} & \multicolumn{4}{ c  }{CityScapes}\\
  &  \multicolumn{2}{ c }{FAE} & \multicolumn{2}{ c | }{BWT} & \multicolumn{2}{ c  }{FAE} & \multicolumn{2}{ c | }{BWT} & \multicolumn{2}{ c }{FAE}  & \multicolumn{2}{ c  }{BWT} \\
  & EPE$\downarrow$ & D1$\downarrow$ & EPE$\downarrow$ & D1$\downarrow$  & EPE$\downarrow$  & D1$\downarrow$ & EPE$\downarrow$ & D1$\downarrow$  & EPE$\downarrow$ & D1$\downarrow$ & EPE$\downarrow$  & D1$\downarrow$ \\
\midrule
Incremental Finetuning & 1.935 & 10.52\% & -0.235 &  1.19\%    & 0.869 & 1.62\% & 0.006 & -0.10\%   & 2.133 & 8.67\% & 0.001 & -0.10\%  \\
EWC~\cite{kirkpatrick2017overcoming} &  1.777 & 8.76\% & -0.121 & -0.78\% & 0.887 & 1.61\% & 0.038 & 0.01\% & 2.252 & 9.22\% & 0.069 & 0.40\%   \\
iCaRL~\cite{rebuffi2017icarl} & 1.799 & \textcolor{blue}{8.34\%} & -0.203 & -0.24\% & \textcolor{red}{0.844} & 1.57\% &  0.016 & -0.06\% & 2.118 & 8.60\% & 0.001 & 0.08\%  \\
Expert Gate~\cite{aljundi2017expert} & 1.895 & 8.61\% & 0.0 & 0.0\%  & 0.877 & 1.75\% & 0.0 & 0.0\%  & 2.181 & 8.63\% & 0.0 & 0.0\% \\
Learn to Grow~\cite{li2019learn} & \textcolor{blue}{1.748} & 8.66\% & 0.0 & 0.0\%  & 0.917 & 1.78\% & 0.0 & 0.0\%  & 2.237 & 8.99\% & 0.0 & 0.0\% \\
Joint Training (Ideal)  & 1.843 & 9.96\% & - & -     & \textcolor{blue}{0.854} & \textcolor{red}{1.52\%} & - & -  & \textcolor{red}{2.105} & \textcolor{red}{8.56\%} & - & -\\
\midrule
RAG (Ours) & \textcolor{red}{1.663} & \textcolor{red}{8.11\%} & 0.0 & 0.0\% & \textcolor{blue}{0.854} & \textcolor{blue}{1.56\%}  &  0.0 & 0.0\%  & \textcolor{blue}{2.111} & \textcolor{blue}{8.57\%} & 0.0 & 0.0\% \\
\bottomrule
\end{tabular}}
\end{center}
\end{table*}

\textbf{Comparisons with MoE.} The Mixture of Experts (MoE) model~\cite{ShazeerMMDLHD17} contains several neural networks as experts with the same architectures, corresponding to potentially different scenes, whose design is similar to the proposed RAG framework. Our approach searches cells at the network level while MoE works at higher levels. Also, MoE needs to calculate which experts to use, which shares similar ideas with the proposed Scene Router.
To this end, we adapt MoE to the stereo matching task for joint training as a baseline for ablation.

\begin{table}[!htbp]
\begin{center}
\caption{Results of cross-dataset comparisons for self-supervised continual stereo.}
\label{cross_dataset_self}
\small
\setlength{\tabcolsep}{1.8mm}{
\begin{tabular}{ c  | c  c | c  c   }
\toprule
{\multirow{2}{*}{Methods}}  &  \multicolumn{2}{ c| }{FAE} & \multicolumn{2}{ c  }{BWT} \\
  & EPE$\downarrow$ & D1$\downarrow$ & EPE$\downarrow$ & D1$\downarrow$  \\
\midrule
Incremental Finetuning & 1.183 & 4.94\% & 0.043 & 0.18\%  \\
EWC~\cite{kirkpatrick2017overcoming} & 1.219 & 5.14\% & 0.072 & 0.67\%    \\
iCaRL~\cite{rebuffi2017icarl}  & 1.217 & 5.31\% & 0.112 & 0.64\%   \\
Expert Gate~\cite{aljundi2017expert} & \textcolor{red}{1.147} & 4.94\% & 0.0 & 0.0\%   \\
Learn to Grow~\cite{li2019learn} &  1.180 &  \textcolor{blue}{4.89\%} & 0.0 & 0.0\%  \\
Joint Training (Ideal)  & 1.193 & 5.28\% & - & - \\
\midrule
RAG (Ours) & \textcolor{blue}{1.154} & \textcolor{red}{4.78\%} & 0.0 & 0.0\%  \\
\bottomrule
\end{tabular}}
\end{center}
\end{table}

Table~\ref{moe} shows the comparison results with the MoE model using four experts on the DrivingStereo dataset. We also include Expert Gate and joint training in the comparison. Although both Expert Gate and MoE are composed of expert models with the same number of tasks, MoE can integrate the predictions of different experts at inference. Thus, its performance is significantly better than Expert Gate. MoE is also superior to joint training because it has multiple expert models to make decisions instead of using a single model. Notably, MoE achieves comparable performance to our RAG framework with Scene Router, which indicates that the function of the sparse gate of MoE and the proposed Scene Router is comparable. Nevertheless, 
our approach has fewer overall model parameters, benefiting from the proposed reusable architecture growth strategy. In contrast, MoE has four times as many overall model parameters as a single expert model.

\subsection{Self-supervised Continual Stereo Evaluation}
\subsubsection{Cross-Scene Comparisons}
Table~\ref{self-incremental} shows the cross-scene comparison results on three real-world datasets for practical self-supervised continual stereo. We can draw several interesting conclusions, some of which are opposite to those in the supervised continual stereo evaluation.

First and foremost, our method achieves compelling performance among all comparison methods on all three datasets. More concretely, we achieve the best results on DrivingStereo and the second-best results on KITTI raw and CityScapes datasets despite being slightly inferior to joint training. This suggests that our method has good scalability and robustness for self-supervised continual stereo. 

Secondly, we observe that regularization and rehearsal-based methods can achieve negative BWT, even the incremental finetuning. This indicates that, under self-supervised conditions, these methods can not only overcome forgetting but also in turn promote previously learned scenes after learning new scenes. Similar observations have been described in~\cite{madaan2022} that unsupervised or self-supervised continual learning can often learn better feature representations to avoid forgetting. Therefore, iCaRL surprisingly yields better performance than Expert Gate on DrivingStereo and KITTI raw, which tends to obtain poor performance under supervised conditions. 

\begin{figure}[t]
\begin{center}
\includegraphics[height=3.01cm,width=8.5cm]{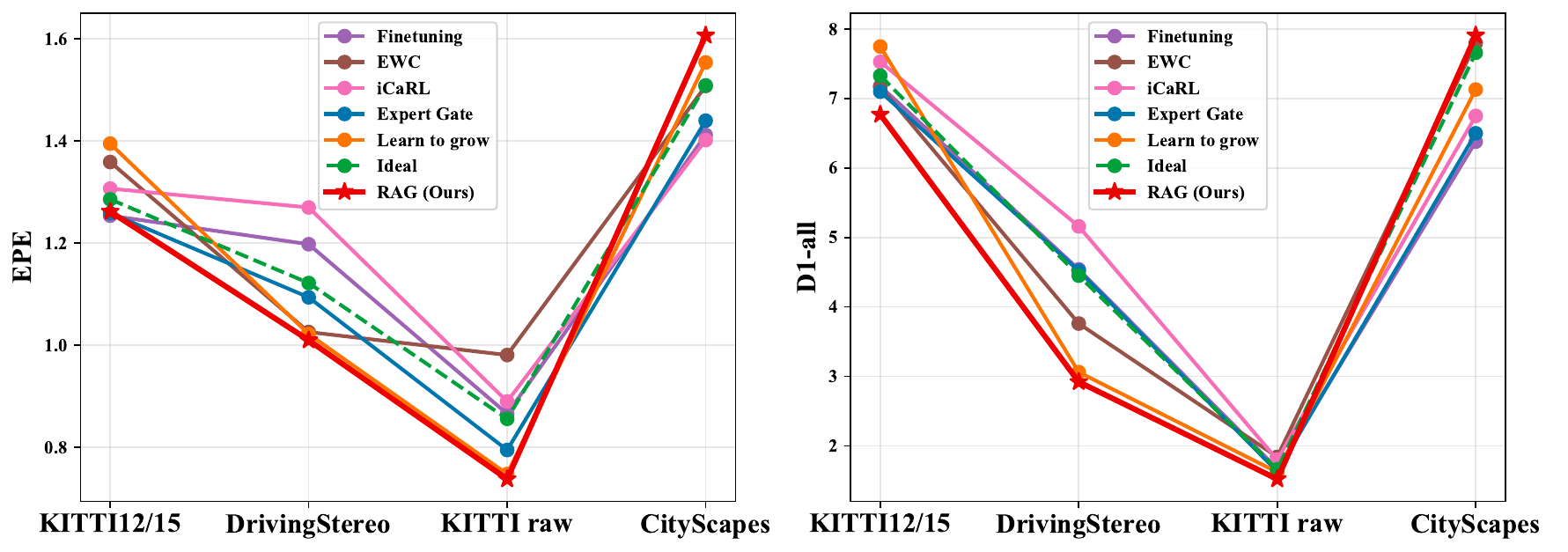}
\end{center}
   \caption{Comparison with comparable baselines on each scene of different datasets for self-supervised continual learning.}
\label{crossdata_self_plot}
\end{figure}

Thirdly, joint training is still a strong baseline on KITTI raw and CityScapes datasets except for DrivingStereo. We speculate that the advantage of joint training on mixed scenes depends on the correlation of each scene. Joint training is promotional for incremental scenes with strong correlation, such as various cities in CityScapes, but not for scenes with weak correlation, like different weather in DrivingStereo.

\begin{table*}[!htbp]
\begin{center}
\caption{Ablation study of neural unit search and architecture growth on the DrivingStereo dataset.}
\label{syn-pretrain}
\small
\setlength{\tabcolsep}{1.8mm}{
\begin{tabular}{  c   | c  c  |   c  c  | c  c  |  c c  |  c c }
\toprule
{\multirow{2}{*}{Substitute subset}}  & \multicolumn{2}{ c | }{Cloudy} &  \multicolumn{2}{ c|  }{Foggy} &  \multicolumn{2}{ c| }{Rainy}  &   \multicolumn{2}{ c | }{Sunny}  &  \multicolumn{2}{ c  }{FAE}  \\
  & EPE $\downarrow$ & D1 $\downarrow$ & EPE $\downarrow$ & D1 $\downarrow$  & EPE $\downarrow$ & D1 $\downarrow$  & EPE $\downarrow$ & D1 $\downarrow$ & EPE $\downarrow$ & D1 $\downarrow$\\
\midrule
None  & 1.083 & 4.39\% & \textbf{1.540} & \textbf{7.44\%} & 3.872 & 18.31\% & 1.086 & 4.30\% & 1.895 & 8.61\%  \\
\midrule
SceneFlow & 1.190 & 5.22\% & 1.634 & 8.22\% & 3.426 & 17.52\% & 1.193 & 5.23\% & 1.861 & 9.05\% \\ 
Flyingthings3D & 1.252 & 6.12\% & 1.896 & 10.31\% & 3.325 & 17.15\% & 1.316 & 5.71\% & 1.947 & 9.82\%   \\
Monkaa & 1.268 & 5.93\% & 1.840 & 10.15\%  & \textbf{2.743} & 17.00\%  & 1.320 & 6.80\% & 1.793 & 9.97\%   \\
Driving & \textbf{1.053} &  \textbf{4.06\%} & 1.659 & 8.10\% & 3.104 & 16.95\% & \textbf{1.021} & \textbf{4.19\%} & \textbf{1.709} & \textbf{8.33\%}    \\
Virtual KITTI & 1.278 & 6.58\% & 1.904 & 9.88\% & 3.151 & \textbf{15.83\%} & 1.260 & 5.17\% & 1.989 & 9.37\%  \\
\bottomrule
\end{tabular}}
\end{center}
\end{table*}

\subsubsection{Cross-dataset Comparisons}
Table~\ref{cross_dataset_self} shows the overall results of cross-dataset comparisons. It is clear that our method achieves the best performance among all the baselines. Notably, joint training does not exhibit a clear advantage due to the large differences in heterogeneous scenes across datasets. Under the self-supervised condition, EWC and iCaRL still achieve low BWT compared to supervised continual stereo. This again shows the good ability of self-supervised learned feature representations to overcome forgetting. Nevertheless, they are inferior to dynamic architecture based  methods since the error accumulation caused by intermediate scenes affects the subsequent model learning. The dynamically growing network can utilize the prior knowledge of previously learned scenes to learn new scenes better, thus avoiding this error accumulation. We further demonstrate the performance comparisons on each scene in Fig.~\ref{crossdata_self_plot}. Our proposed RAG achieves the best performance in three out of four scenes benefitting from scene-specific neural unit search and reusable network growth.

\begin{table}[t]
\caption{Ablation results of different strategies for the model reusability of self-supervised continual stereo on the DrivingStereo dataset.}
\small
\subfloat[Comparison results using different initialization values. \label{tab_unsa}]{
\setlength{\tabcolsep}{4mm}{
\begin{tabular}{l | c c c }
\toprule
Initial Value & EPE $\downarrow$ & D1 $\downarrow$ & ARR $\uparrow$\\
\midrule
$c_0 = 0$ & 1.937 & 10.47\% & \textbf{68.5\%}  \\
$c_0 = 5$ & 1.911 & \textbf{9.65\%} & 53.7\% \\
$c_0 = 10$ & \textbf{1.905} & 10.02\% & 29.6\% \\
\bottomrule
\end{tabular}}}\hspace{3mm}
\subfloat[Comparison results using different probability multiples. \label{tab_unsb}]{
\setlength{\tabcolsep}{4.5mm}{
\begin{tabular}{l | c c c}
\toprule
Multiple & EPE $\downarrow$ & D1 $\downarrow$ & ARR $\uparrow$\\
\midrule
$\gamma = 1$ & 1.911 & 9.65\% & 53.7\%  \\
$\gamma = 2$ &  \textbf{1.834} & \textbf{9.34\%} & \textbf{57.4\%} \\
$\gamma = 3$ & 1.860 & 10.22\% & \textbf{57.4\%} \\
\bottomrule
\end{tabular}}}\hspace{3mm}
\subfloat[Comparison results using different validation scores. \label{tab_unsc}]{
\setlength{\tabcolsep}{3.2mm}{
\begin{tabular}{c | c c c}
\toprule
Validation Score & EPE $\downarrow$ & D1 $\downarrow$ & ARR $\uparrow$\\
\midrule
Eq.~\eqref{eq:simple} & 1.834 & 9.34\% & \textbf{57.4\%}  \\
Eq.~\eqref{eq:alter} & 2.174 & 10.53\% & 46.3\%  \\
Eq.~\eqref{eq:ours} & \textbf{1.813} & \textbf{9.08\%} & 50.1\%  \\
\bottomrule
\end{tabular}}}\hspace{3mm}
\subfloat[Comparison results using different target parameters. \label{tab_unsd}]{
\setlength{\tabcolsep}{3mm}{
\begin{tabular}{c | c c c}
\toprule
Target Parameters & EPE $\downarrow$ & D1 $\downarrow$ & ARR $\uparrow$\\
\midrule
$\phi = \Phi/3$ & 1.937 & 10.07\% & 37.8\%  \\
$\phi = \Phi/2$ & 1.813 & 9.08\% & 50.1\% \\
$\phi = \Phi$ & \textbf{1.794} & \textbf{8.96\%} & \textbf{50.4\%} \\
\bottomrule
\end{tabular}}}\hspace{3mm}
\label{tab_uns}
\end{table}

\subsubsection{Ablation Study}
\label{proxy}

\textbf{Proxy-supervision.}
In self-supervised continual stereo, we propose the proxy-supervised architecture growth strategy to alleviate the challenge of the lack of disparity labels. Table~\ref{syn-pretrain} shows the comparison results of different synthetic datasets or subsets for neural unit search on each scene. Using Monkaa and Flyingthings3D subsets as proxy-supervision is inferior to directly using the structure searched on synthetic data. This is because their content is cartoon characters or flying objects, which are quite different from driving scenarios. In contrast, adopting the Driving subset can achieve better performance in three of the total four scenes since they are simulated driving scenarios, which are even better than adopting the Scene Flow dataset. This suggests that using content-irrelevant synthetic data is redundant. Notably, adopting the Virtual KITTI datasets does not obtain expected good performance, although it also contains driving scenes. We speculate that various weather conditions are simulated in the dataset resulting in low discrimination of the transferred images for each scene. The above analysis verifies the effectiveness of our proxy-supervision strategy of replacing real-world data with transferred synthetic driving data.

\textbf{Reusability.}
Akin to supervised continual stereo, we also explore the effect of the four factors on the self-supervised continual stereo as shown in Table~\ref{tab_uns}.
In Table~\ref{tab_unsa}, adopting a non-zero initialization can achieve better performance despite slightly lower reuse rates. To balance the performance and model reusability, we finally choose $c_0 = 5$ for proxy-supervised architecture growth. Similar to the supervised architecture growth, Table~\ref{tab_unsb} shows that using $\gamma = 2$ reduces the error rates and improves the model reuse rates, which achieves the best performance. In Table~\ref{tab_unsc}, compared to ordinary validation scores in Eq.~\eqref{eq:simple} and Eq.~\eqref{eq:alter}, our designed strategy in Eq.~\eqref{eq:ours} still achieves the best trade-off between performance and reuse rate in self-supervised continual stereo. Different from supervised architecture growth, Table~\ref{tab_unsd} shows that using larger target weight values can achieve better performance. Thus we finally choose $\phi=\Phi$.

\section{Discussion}
\label{discussion}
\subsection{Adaptation to Unseen Scenes}
We further compare our method with continuous adaptation methods~\cite{MADNet,poggi2021continual} to show the ability to overcome forgetting and the adaptability to unseen scenes at inference. The version called MADNet-GT-Full is adopted for supervised continual stereo comparison while MAD++ is adopted for comparison in a self-supervised manner. Both methods learn or adapt to a series of scenes on the DrivingStereo dataset, $i.e.$, \emph{cloudy} $\rightarrow$ \emph{foggy} $\rightarrow$ \emph{rainy} $\rightarrow$ \emph{sunny}.

\begin{figure}[t]
\begin{center}
\includegraphics[height=7.66cm,width=8.5cm]{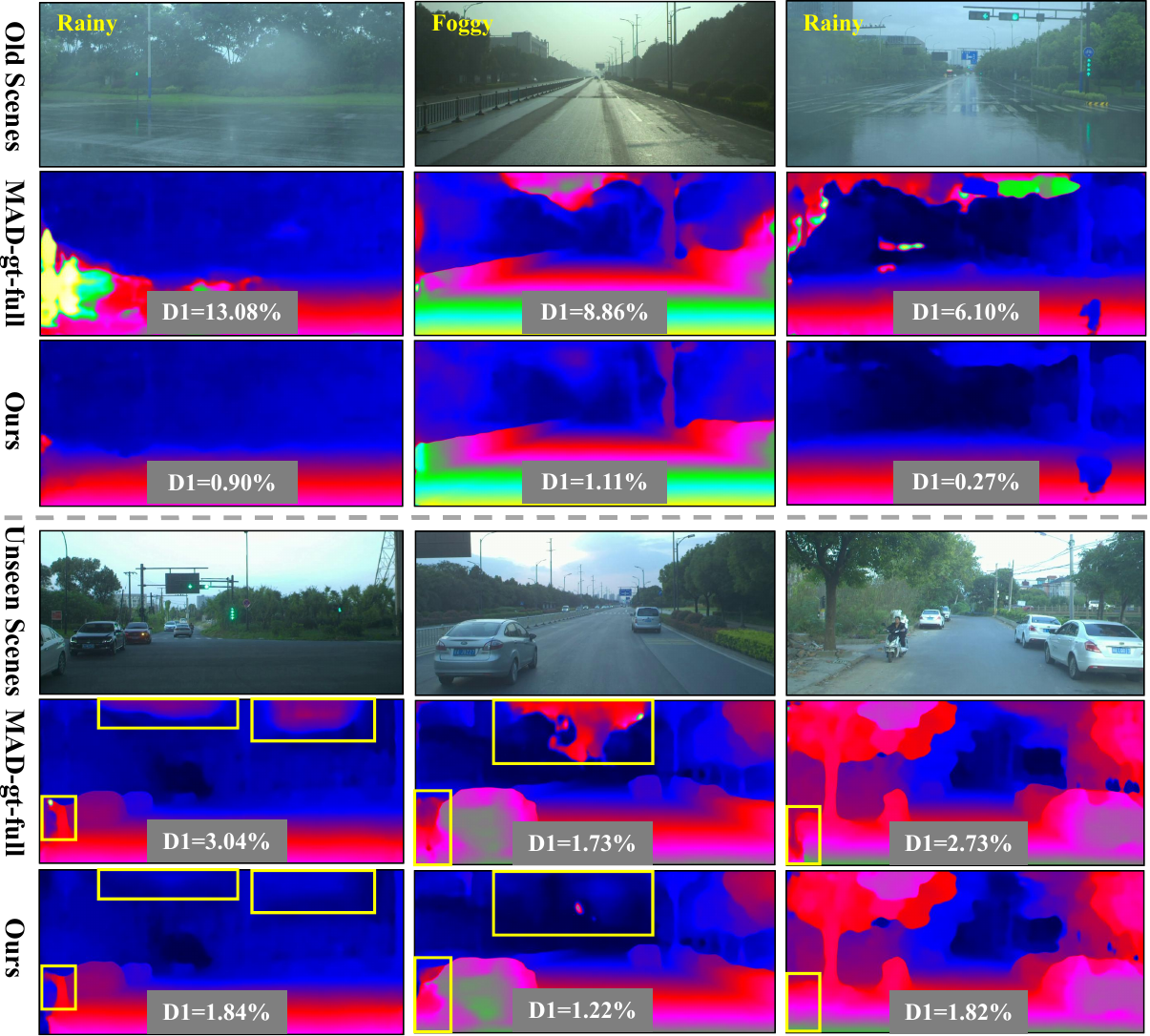}
\end{center}
   \caption{Visual comparisons of disparity map with MAD-gt-full~\cite{MADNet} for previously learned scenes (top three rows) and unseen scenes (bottom three rows). The yellow box marks areas that are significantly improved.}
\label{madgt}
\end{figure}

For the supervised condition, as shown in Fig.~\ref{madgt}, MAD-GT-Full yields severe errors when tested on previously learned scenes ($e.g.,$ \emph{rainy} or \emph{foggy} days) since it has forgotten previous knowledge. It needs some buffer time to re-adapt to this scene through continuous online gradient updates. In contrast, our method can overcome catastrophic forgetting to achieve good results. Benefiting from the Scene Router, our method can quickly adapt to the rapid scene switches. We also show the generalization performance of direct inference without adaptation in the bottom rows. When exposed to an unseen scene, such as \emph{overcast} days at dusk, our Scene Router can adaptively select the suitable architecture path to predict better disparity. In the three samples, the architecture corresponding to \emph{cloudy}, \emph{foggy}, and \emph{rainy} scenes are selected for inference, respectively, since the scene type of input images are closest to these days.

\begin{figure}[t]
\begin{center}
\includegraphics[height=7.68cm,width=8.5cm]{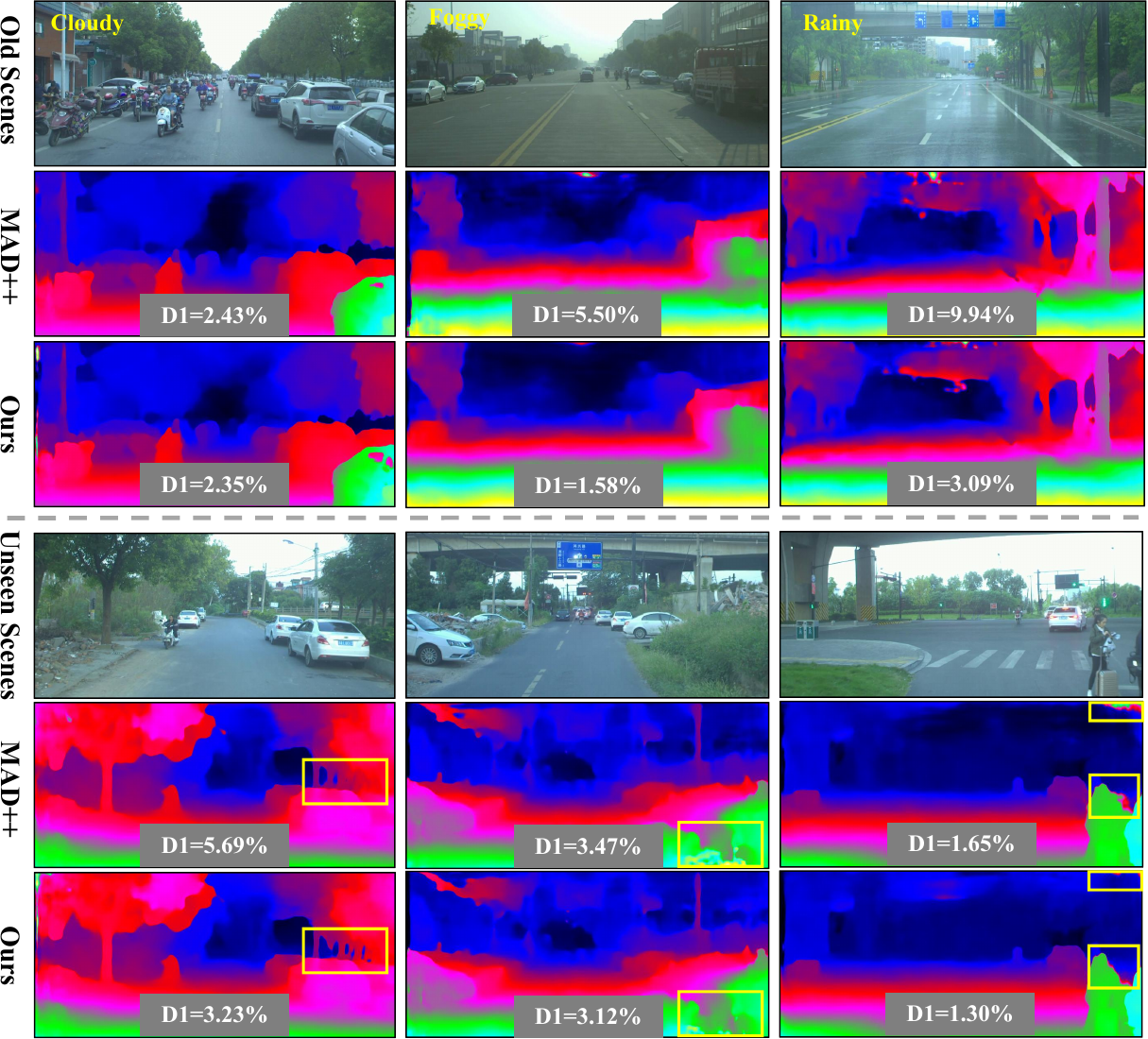}
\end{center}
   \caption{Visual comparisons of disparity map with MAD++~\cite{poggi2021continual} for previously learned scenes (top three rows) and unseen scenes (bottom three rows). The yellow box marks improved areas.}
\label{mad_plus}
\end{figure}

Fig.~\ref{mad_plus} shows the cases for the self-supervised condition. MAD++ can alleviate forgetting by utilizing SGM proxy labels and uncertainty filtering in most scenes like \emph{cloudy} and \emph{foggy}, but it still suffer catastrophic forgetting on the challenge \emph{rainy} scene. In contrast, our method can adaptively select suitable scene-specific architectures for inference, thus overcoming forgetting. Besides, our method also generalizes better to unseen scenes. In the three examples on the bottom, the \emph{cloudy} path is chosen at inference according to the scene type of input images, while MAD++ can only rely on the lately learned \emph{sunny} scene.

\subsection{Architecture Growth Analysis}
\begin{figure}[t]
\begin{center}
\includegraphics[height=5.04cm,width=8cm]{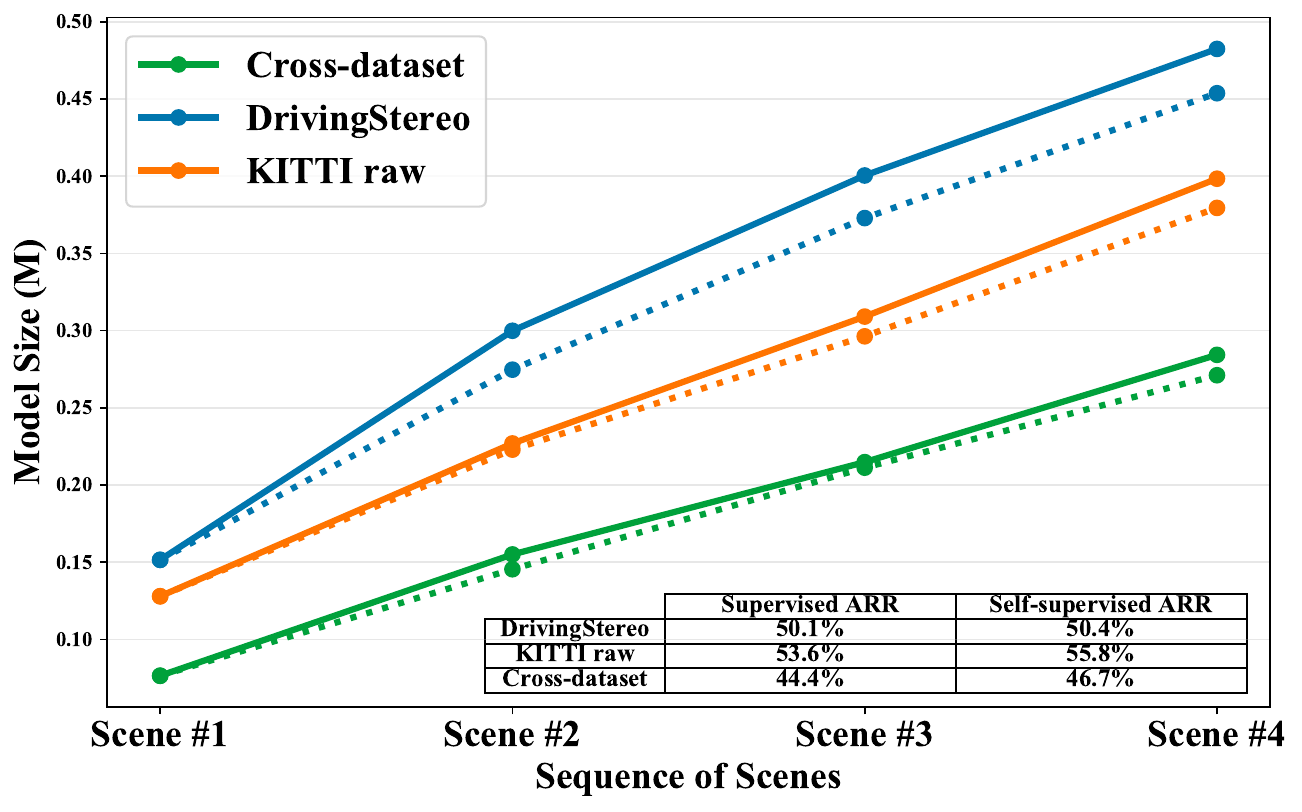}
\end{center}
   \caption{Comparison of architecture growth trends and average reuse rate (ARR) on various datasets under supervised (solid line) and self-supervised (dashed line) conditions.}
\label{model_grow}
\end{figure}

To further explore the reusability of the proposed architecture growth strategy in different environments, we depict the model growth trend as shown in Fig.~\ref{model_grow}. We observe that the model grows the slowest on the KITTI raw dataset, the fastest on the cross-dataset conditions, and moderately on the DrivingStereo dataset. Correspondingly, the model has the highest reuse rate on KITTI raw and the lowest reuse rate under cross-dataset conditions. This is because the correlation between the scenes on KITTI raw is relatively high thus previously learned neural units can better promote the learning of subsequent scenes. 
Conversely, under the challenging cross-dataset conditions, different scenes in various datasets have significant differences. Thus the model needs more new neural units to learn new scenes, which decreases the number of reused neural units. It is applicable for both supervised and self-supervised continual stereo. Notably, compared with the supervised condition, the model grows slower, and the reusability is generally higher for the self-supervised condition. This further verifies our previous analysis that self-supervised learned feature representation is more capable of overcoming forgetting.

\subsection{Task Order}
\subsubsection{Easy-hard Order}
\label{sec:easy_hard}
In the previous experiments, we used a randomly chosen task order to explore continual stereo. Under this general setting, we observe that the difficulty of learning various scenes differs for both supervised and self-supervised conditions. For example, rainy and foggy scenes are easier to learn in supervised conditions but hard to learn in self-supervised conditions. Here, the difficulty of learning refers to the performance that a single model can reach on a single scene. \emph{So what is the relationship between the learning order of the scenes and the final performance?} To answer the question, we further choose two intentionally designed orders named Easy Order and Hard Order for experiments. The Easy Order means learning the easiest task first and then more difficult tasks, which are \emph{foggy} $\rightarrow$ \emph{rainy} $\rightarrow$ \emph{cloudy} $\rightarrow$ \emph{sunny} for supervised continual stereo and \emph{sunny} $\rightarrow$ \emph{cloudy} $\rightarrow$ \emph{foggy} $\rightarrow$ \emph{rainy} for self-supervised continual stereo. The Hard Order is the reverse order. 

\begin{figure}[t]
\begin{center}
\includegraphics[height=6.11cm,width=8.5cm]{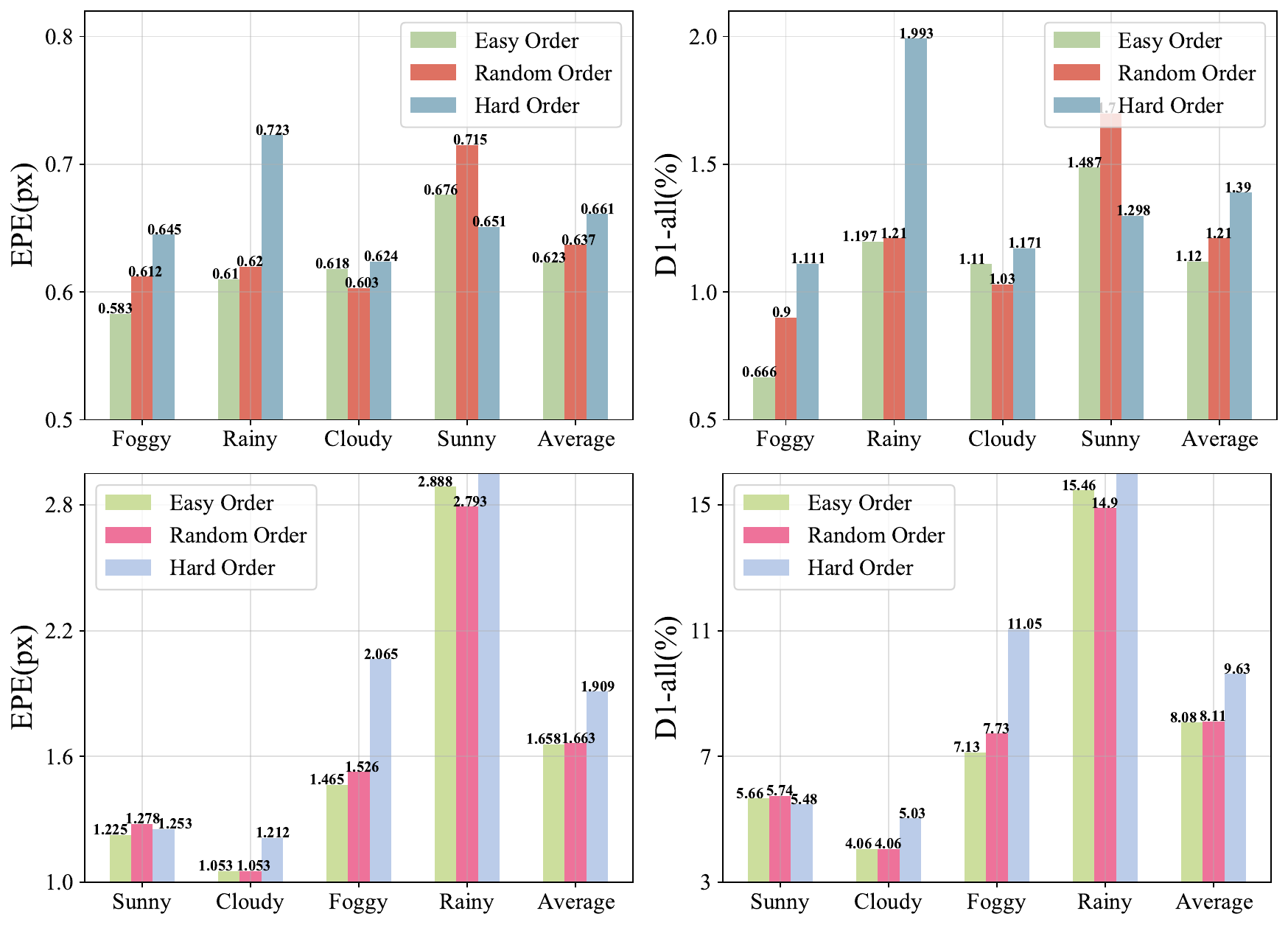}
\end{center}
   \caption{Comparison of the results of different task orders on the DrivingStereo dataset for supervised (top two) and self-supervised (bottom two) continual stereo, which are random order, easy order, and hard order for the respective situation.}
\label{task-order}
\end{figure}

Fig.~\ref{task-order} depicts the comparison results for both conditions. We observe that learning easy tasks and then difficult tasks achieves the best performance in the end and vice versa. This suggests that learning simple tasks lays the foundation and provides prior knowledge for promoting subsequent complex tasks, thus making them easier to learn. It is similar to how humans acquire knowledge since we can understand the more difficult calculus only if we first learn elementary arithmetic.

\subsubsection{Repetition Order}
The four weather sequences were not repeated in the previous experiments. In practice, it is possible to encounter newly collected data that belongs to the same domain as the previously learned scenes. To explore the results under repetitive scenes, we split each weather scene into sub-scenes and learning in the order of $cloudy_1(C_1)$ $\rightarrow$ $foggy_1(F_1)$ $\rightarrow$ $rainy_1(R_1)$ $\rightarrow$ $sunny_1(S_1)$ $\rightarrow$ $cloudy_2(C_2)$ $\rightarrow$ $foggy_2(F_2)$ $\rightarrow$ $rainy_2(R_2)$ $\rightarrow$ $sunny_2(S_2)$.

\begin{figure}[t]
\begin{center}
\includegraphics[height=6.28cm,width=8.5cm]{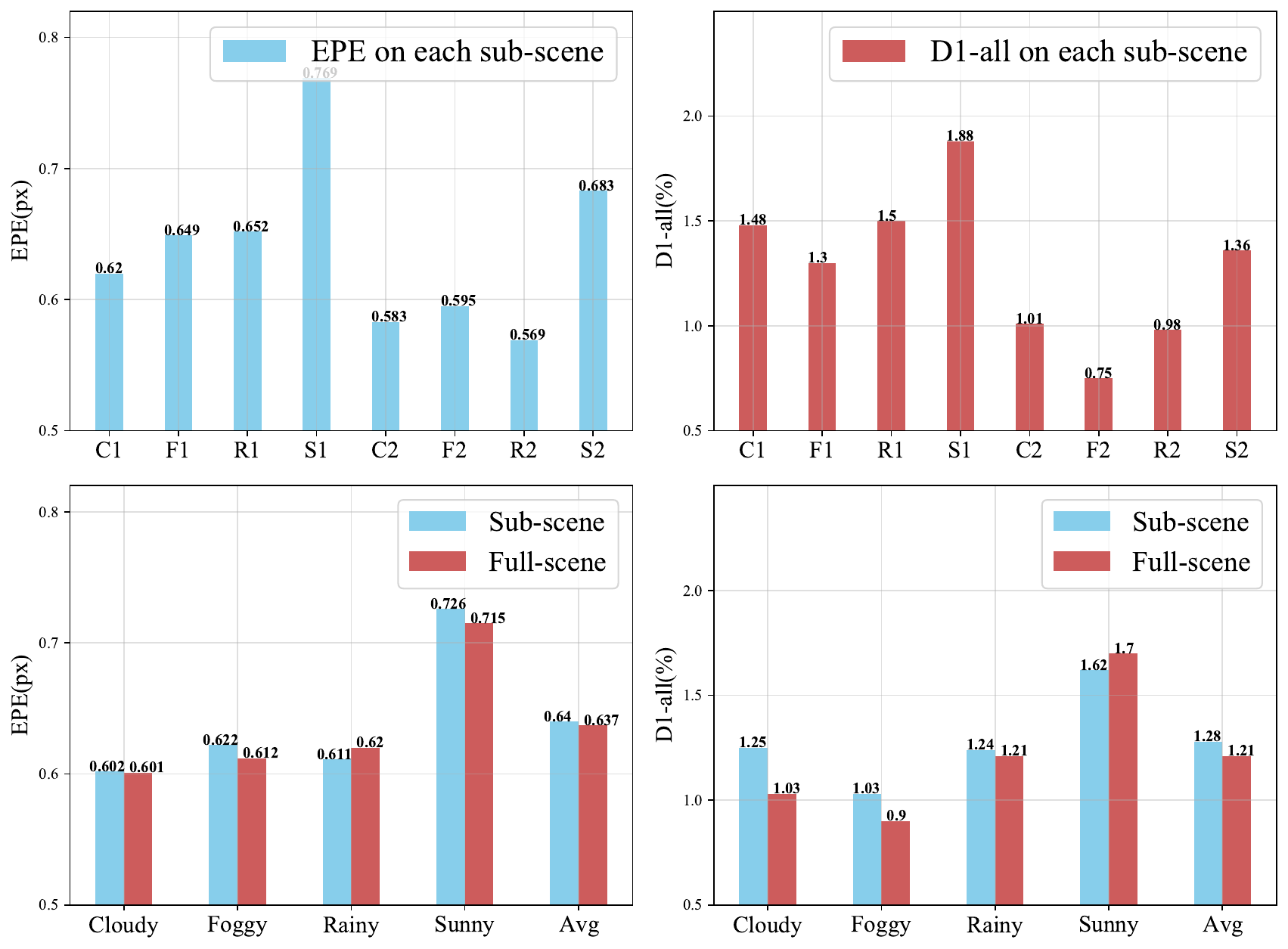}
\end{center}
   \caption{Specific results under supervised continual stereo in each sub-scene (top row) and the comparison between full-scene and sub-scene orders (bottom row).}
\label{repetitive_order}
\end{figure}

\begin{table}[!htbp]
\begin{center}
\caption{Average results of cross-dataset comparisons using all sequences possible with four datasets for self-supervised continual stereo.}
\label{cross_dataset_self_24}
\small
\setlength{\tabcolsep}{1.8mm}{
\begin{tabular}{ c  | c  c | c  c   }
\toprule
{\multirow{2}{*}{Methods}}  &  \multicolumn{2}{ c| }{FAE} & \multicolumn{2}{ c  }{BWT} \\
  & EPE$\downarrow$ & D1$\downarrow$ & EPE$\downarrow$ & D1$\downarrow$  \\
\midrule
Incremental Finetuning & 1.169 & 4.76\% & 0.039 & 0.15\%  \\
EWC~\cite{kirkpatrick2017overcoming} & 1.201 & 5.02\% & 0.067 & 0.62\%    \\
iCaRL~\cite{rebuffi2017icarl}  & 1.198 & 5.17\% & 0.101 & 0.60\%   \\
Expert Gate~\cite{aljundi2017expert} & \textcolor{blue}{1.146} & 4.92\% & 0.0 & 0.0\%   \\
Learn to Grow~\cite{li2019learn} & 1.168  &  \textcolor{blue}{4.75\%} & 0.0 & 0.0\%  \\
Joint Training (Ideal)  & 1.193 & 5.28\% & - & - \\
\midrule
RAG (Ours) & \textcolor{red}{1.145} & \textcolor{red}{4.69\%} & 0.0 & 0.0\%  \\
\bottomrule
\end{tabular}}
\end{center}
\end{table}

Fig.~\ref{repetitive_order} shows the specific results under supervised continual stereo in each sub-scene (top row) and the comparison results between full-scene and sub-scene orders (bottom row). Three interesting things can be observed. i) The performance in the four sub-scenes in the first stage is inferior to that in the full scene since the training data is reduced to half of the original. ii) Regarding the second stage, the performance of the four sub-scenes is better than that of the corresponding scenes in the first stage and even superior to that of the full scene. It seems that the repetitive scenes in the second stage enable our model to reuse the neural units learned from the previous similar scenes, thus achieving knowledge forward transfer. iii) Learning with sub-scene and full-scene orders achieves comparable average performance, yet learning with the full-scene order is slightly better. We speculate that splitting each weather scene into sub-scenes increases the number of tasks, thus causing the model not to learn as well as before.

\subsubsection{Permutation Order}
In addition to exploring task orders in terms of difficulty and repetition, we further explore all possible task sequences with four datasets under cross-dataset evaluations. They are KITTI 2012/2015, DrivingStereo (\emph{cloudy}), KITTI raw (\emph{campus}), and CityScapes (\emph{bremen}) with a total of 24 possible orders obtained by permutation. 

\begin{table*}[!htbp]
\begin{center}
\caption{Monocular depth estimation results of cross-scene comparisons for supervised continual evaluations on the DrivingStereo dataset.}
\label{monocular_depth}
\normalsize 
\setlength{\tabcolsep}{4mm}{
\begin{tabular}{ c  | c c c | c  c  c }
\toprule
Methods & RMSE$\downarrow$ & REL$\downarrow$ & Log10$\downarrow$ & $\delta_1\uparrow$ & $\delta_2\uparrow$ & $\delta_3\uparrow$ \\
\midrule
Incremental Finetuning & 0.192 & 10.235 & 0.097 & 0.679 & 0.863 & 0.944   \\
Expert Gate~\cite{aljundi2017expert} & 0.178 & 9.701 & 0.084 & 0.725 & 0.888 & \textcolor{blue}{0.961} \\
Learn to Grow~\cite{li2019learn} & 0.174 & 9.652 & 0.083 & 0.721 & 0.885 & 0.956    \\
Joint Training (Ideal)  & \textcolor{red}{0.164} & \textcolor{blue}{9.637} & \textcolor{red}{0.069} & \textcolor{red}{0.723} & \textcolor{red}{0.902} & \textcolor{red}{0.967} \\
\midrule
RAG (Ours) & \textcolor{blue}{0.170} & \textcolor{red}{9.626} & \textcolor{blue}{0.077} & \textcolor{blue}{0.716} & \textcolor{blue}{0.896} & \textcolor{red}{0.967} \\
\bottomrule
\end{tabular}}
\end{center}
\end{table*}

Table~\ref{cross_dataset_self_24} shows the average results across all the sequences for the cross-dataset evaluations. Changing the sequence of scenes does not affect joint training, and it has little effect on Expert Gate since it assigns a separate model to each scene. We are surprised that the continual learning methods perform better than learning in the random order in Table~\ref{cross_dataset_self}. It seems to be explained by the conclusions we explored in Section~\ref{sec:easy_hard}. According to Fig.~\ref{crossdata_self_plot}, KITTI Raw is the easiest dataset to learn, while CityScapes is the most difficult. The random order used in the previous experiment is closer to the most difficult order, that is, learning a hard scene like KITTI 2012/2015 first and then learning an easy scene like KITTI Raw. Thus, the performance is also slightly worse than average. On the contrary, the overall performance is higher since it covers easy and hard scenes in various permutation combinations. Yet, the overall performance gains are modest.

\section{Extension}
\label{extention}
The proposed framework has the potential to work on a broad range of tasks in the same continual setting. To demonstrate the scalability of the proposed approach, we extend the RAG framework to more fundamental tasks like monocular depth estimation and high-level 3D tasks like stereo-based 3D object detection.

\subsection{Monocular Depth Estimation}
We make the following modifications to adapt the RAG framework to the monocular depth estimation task.

\textbf{Network Structure.} We first perform feature extraction only on the left image and remove the cost volume from stereo matching. Then, operations based on the 3D convolution in the Matching Net are modified to 2D convolution-based operations to adapt the Matching Net to the monocular depth estimation model as a decoder. We finally change the disparity regression head to a depth regression head following NewCRFs~\cite{yuan2022newcrfs}.

\textbf{Datasets and Metrics.} Four weather scenes of the DrivingStereo dataset and the corresponding depth labels are used for model training and evaluation. We use standard evaluation metrics following previous work~\cite{yuan2022newcrfs}: 1) Root mean square error (RMSE); 2) Mean absolute relative error (REL); 3) Mean log10 error (log10); 4) Accuracy under threshold $t$ ($t \in [1.25, 1.25^2, 1.25^3]$).

Table~\ref{monocular_depth} shows the cross-scene evaluation results of supervised continual monocular depth estimation on the DrivingStereo dataset. It can be seen that the monocular depth estimation model also suffers from domain shifts between different weather scenes since the average performance of the incremental finetuning method in each scene is poor. Under this challenge, the proposed method shows good scalability in the monocular depth estimation task, significantly outperforming other methods based on parameter isolation. It is worth noting that our method is slightly inferior to joint training on some indicators, which suggests that the monocular depth estimation model benefits more from training on diverse mixed data.

\begin{table}[!htbp]
\begin{center}
\caption{Stereo-based 3D object detection results of cross-dataset comparisons for supervised continual evaluations on the KITTI validation set.}
\label{3d_detection}
\small
\setlength{\tabcolsep}{2.5mm}{
\begin{tabular}{ c  | c  c  c    }
\toprule
{\multirow{2}{*}{Methods}}  &  \multicolumn{3}{ c }{Car AP$_{3D}$ (IoU=0.7)}  \\
  & Easy & Moderate & Hard \\
\midrule
Incremental Finetuning & 66.79\% & 42.42\% & 31.77\%  \\
Expert Gate~\cite{aljundi2017expert} & 70.28\% & 44.02\% & \textcolor{blue}{33.74\%}  \\
Learn to Grow~\cite{li2019learn} & 70.65\% & 44.62\% & 33.08\%  \\
Joint Training (Ideal) & \textcolor{red}{72.06\%} & \textcolor{red}{46.58\%} & \textcolor{red}{35.53\%} \\
\midrule
RAG (Ours) & \textcolor{blue}{71.26\%} & \textcolor{blue}{45.19\%} & 33.65\%  \\
\bottomrule
\end{tabular}}
\end{center}
\end{table}

\subsection{Stereo-based 3D Object Detection }
We make the following modifications to adapt the RAG
framework to the stereo-based 3D object detection task.

\textbf{Network Structure.}
We adopt YOLOStereo3D~\cite{LiuWL21} as the stereo-based 3D detector since it is close to the stereo matching framework. Specifically, the four blocks of the ResNet-34 backbone, the convolution layers of Ghost Module, and the convolution layers of multi-scale fusion are defined as cells. We assign separate box regression and classification heads and disparity estimation heads to each task. As the number of tasks increases, the 3D detector grows in a reusable manner according to the proposed network-level growth strategy in Section~\ref{sec:network_grow}.

\textbf{Datasets and Metrics.}
Following YOLOStereo3D~\cite{LiuWL21}, we adopt the KITTI~\cite{kitti12} dataset for stereo-based 3D object detection. We split the training data into a training set with 3712 samples and a validation set with 3769 samples. We further divide the training set into three parts with equal number of samples and name them $KITTI_1$, $KITTI_2$, and 
$KITTI_3$ to build a continual learning task, whose sequences are $KITTI_1$ $\rightarrow$ $KITTI_2$ $\rightarrow$ $KITTI_3$. Objects are divided into three difficulty regimes: \emph{easy}, \emph{moderate}, and \emph{hard}, according to their 2D box height, occlusion, and truncation levels following the KITTI setting. We report the average precision of 3D boxes (AP$_{3D}$) of \emph{Cars} using IoU=0.7.

Table~\ref{3d_detection} shows the cross-dataset evaluation results of supervised continual stereo based 3D object detection on the KITTI validation set. One can see some domain shifts between the three randomly divided sub-datasets since the result of the incremental finetuning setting yields the worst performance. Joint training achieves the best performance on various object levels, suggesting that increasing training data is a shortcut to improving model performance. The parameter isolation based methods achieve comparable performance, among which the proposed RAG framework obtains higher accuracy on easy and moderate objects, which validates the scalability and advantages of the proposed method for the high-level 3D detection task. Notably, Expert Gate is slightly better on hard objects. The dynamic structure based approach seems to not learn well enough due to few samples of hard objects.

\section{Conclusion and Future Work}
\label{conclusion}
We have presented a reusable architecture growth framework to tackle the continual stereo problem. By leveraging task-specific neural unit search and architecture growth with real-world ground truth disparity or synthetic proxy-supervision, our framework can continually learn new scenes while not forgetting previously learned scenes with high model reusability. At deployment, we utilize Scene Router to adaptively select the architecture path to adapt to rapid scene switches without online gradients update. Experimental results on various datasets support the effectiveness of our proposal, highlighting in particular how our method is more beneficial than various continual learning methods in challenging weather, road, city, and cross-dataset conditions. The experimental findings also provide evidence of the ability of our paradigm to learn knowledge that can generalize well to novel unseen scenes.

In future work, we plan to explore how to make the reused units tuned on the new task while overcoming forgetting. This may significantly reduce the overall model size during growth. We would also like to investigate alternative approaches to realize the function of adaptively selecting architecture paths for the inputs.


%



\ifCLASSOPTIONcompsoc
  \section*{Acknowledgments}
\else
   regular IEEE prefers the singular form
  \section*{Acknowledgment}
\fi

This work was supported partially by the National Natural Science Foundations of China (Grants No.62376267, 62222302, 62076242), the Pre-Research Project on Civil Aerospace Technologies (No. D030312), the National Defense Basic Scientific Research Program of China (No. JCKY2021203B063), and the innoHK project.

\ifCLASSOPTIONcaptionsoff
  \newpage
\fi



%




\bibliographystyle{IEEEtran}
\bibliography{pami_bib}

%




\end{document}